\documentclass[a4paper,fleqn]{cascas}

\usepackage{graphicx}
\usepackage{amssymb}
\usepackage{amsmath}

 \usepackage{float}
\usepackage[numbers]{natbib}
\usepackage[textsize=small,colorinlistoftodos,color=orange]{todonotes}
\usepackage{algorithm2e}
\newcommand\tx[1]{\todo[inline,shadow]{\expandafter\MakeUppercase #1}}
\usepackage{subcaption}
\newcommand\ty[1]{}

\usepackage{stix}
\usepackage{enumitem}
\usepackage{caption}
\usepackage{wrapfig}
\usepackage{adjustbox}
\usepackage{xspace}
\usepackage{soul}
\usepackage{lineno}
\usepackage{tikz}
\usetikzlibrary{positioning}

\usepackage{setspace}

\usepackage{orcidlink}

\title[mode=title]{Predicting the duration of traffic
incidents for Sydney greater metropolitan area using machine learning methods}

\author[uts]{Artur Grigorev\corref{cor1}\orcidlink{0000-0001-6875-3568}}
\ead{Artur.Grigorev@uts.edu.au}
\ead[url]{www.fmlab.org}
\cortext[cor1]{Corresponding author}

\author[uts]{Sajjad Shafiei \orcidlink{0000-0002-0155-6866}}
\ead{sajjad.shafiei@uts.edu.au}

\author[unsw,data61]{Hanna Grzybowska \orcidlink{0000-0003-2614-5964}}
\ead{hanna.grzybowska@data61.csiro.au}

\author[uts]{Adriana-Simona Mih\u{a}i\c{t}\u{a}\corref{cor2} \orcidlink{0000-0001-7670-5777}}
\ead{Adriana-Simona.Mihaita@uts.edu.au}
\ead[url]{https://www.uts.edu.au/staff/adriana-simona.mihaita}

\address[uts]{Faculty of Engineering and IT, University of Technology Sydney, Sydney, Australia}
\address[data61]{Data61, CSIRO, Eveleigh, NSW 2015, Australia}
\address[unsw]{School of Civil and Environmental Engineering, University of New South Wales, Sydney, NSW, Australia}

\shortauthors{Grigorev et~al.}

\shorttitle{Predicting the duration of traffic
incidents for Sydney greater metropolitan area using machine learning methods}

\usepackage{setspace}

\begin{document}


\begin{abstract}
This research presents a comprehensive machine learning approach to predicting the duration of traffic incidents, classifying them as short-term or long-term, and understanding what are the factors that affect the duration the most. Our modelling methodology is using a dataset from the Sydney Metropolitan Area that includes detailed records of traffic incidents, road network characteristics, and socio-economic indicators, we train and evaluate a variety of advanced machine learning models including Gradient Boosted Decision Trees (GBDT), Random Forest, LightGBM, and XGBoost. The models are assessed using Root Mean Square Error (RMSE) for regression tasks and F1 score for classification tasks.

Our experimental results demonstrate that XGBoost and LightGBM outperform conventional models with XGBoost achieving the lowest RMSE of 33.7 for predicting incident duration and highest classification F1 score of 0.62 for a 30-minute duration threshold. For classification, the 30-minute threshold balances performance with 70.84\% short-term duration classification accuracy and 62.72\% long-term duration classification accuracy. Feature importance analysis, employing both tree split counts and SHAP values, identifies the number of affected lanes, traffic volume, and types of primary and secondary vehicles as the most influential features.

The proposed methodology not only achieves high predictive accuracy but also provides stakeholders with vital insights into local area factors contributing to incident durations. These insights enable more informed decision-making for traffic management and response strategies. The code is available by the link: \url{https://github.com/Future-Mobility-Lab/SydneyIncidents}

\end{abstract}



\maketitle

\section{Introduction}

Traffic incidents are a significant concern in Australia, with the country experiencing a high rate of road fatalities and injuries. According to the Australian Road Deaths Database (ARDD) and the National Crash Database (NCD), there were 1,194 road crash deaths in 2022, representing a 5.8\% increase from 2021. The statistics on fatal road crashes in Australia are alarming, with a total of 333 deaths in 2013, increasing to 1,194 deaths in 2022. Over the last decade, national fatalities have remained largely flat, with a slight increase in motorcyclist deaths and a marginal reduction in pedestrian and pedal cyclist deaths. The majority of road deaths occur in Regional and Remote areas, with one-third occurring in Major City areas. In addition to the statistics on fatal road crashes, the Quarterly Bulletin of Serious Injury Crash Data provides insights into the trend of serious injuries. Over the 12-month period ending September 2023 there were 10,021 serious injuries across Australia.

Efficient prediction and management of road traffic incidents are crucial in minimizing congestion and enhancing safety, particularly in densely populated areas. The precision of these predictions profoundly impacts emergency response times and overall traffic management strategies. Central to these predictions is the estimation of 'incident duration'. The term 'incident duration' can refer to different time spans. Some studies define it as the interval from the onset of an incident to the clearance of the scene from the main road by the response team, while others consider it the period until the restoration of normal traffic conditions. The latter definition often requires more detailed analyses of traffic conditions and traveller behaviour. Both durations are influenced by factors such as the nature of the incident, the effectiveness of the emergency responses, and the prevailing road and traffic conditions.

The practical applications of accurate incident duration predictions are extensive. For emergency response units, including ambulance and fire services, these predictions facilitate strategic planning concerning the placement and readiness of teams. Traffic management authorities also use these insights to optimize the number and location of response units, effectively reducing incident durations. Furthermore, integrating real-time traffic simulation or traffic predictive systems with incident duration models can revolutionize traffic management strategies. These integrations allow for dynamic adjustments of traffic signals and route advisories based on ongoing incidents, showcasing the transformative potential of advanced predictive analytics in urban traffic management.
Historical incident data plays a pivotal role in developing predictive models. While some studies focus on exact duration predictions, others classify incident durations into categories such as short-term or long-term, aiding in rapid and effective response strategies. Various models, ranging from simple regression techniques to sophisticated machine learning algorithms like Gradient Boosted Decision Trees, Random Forest, and XGBoost, are employed to estimate incident durations. These models not only predict the time needed to resolve incidents but also assist in resource and response planning by classifying incident by durations appropriately.

The effectiveness of these models largely depends on the granularity and relevance of the data used. Recent compilations of detailed road incident data, particularly for the Sydney metropolitan area, offer a unique opportunity to refine these models. This study represents one of the pioneering efforts to apply state-of-the-art machine learning techniques to such comprehensive datasets, aiming to set new benchmarks in the accuracy and reliability of incident duration predictions. By focusing on Local Government Area (LGA) data within the Sydney Metropolitan Area, this study explores the relationship between incident duration and local road parameters, economic and travel behaviour characteristics. Leveraging this detailed dataset, we aim to provide insights that enhance the development of predictive models, thereby improving traffic management strategies with the consideration of these variables.

This local focus is crucial as it acknowledges the specific traffic distribution, demand, and road network features of Sydney, addressing a critical gap in existing research and capturing the unique characteristics of road traffic incidents that can vary significantly from one local area to another.

Despite the availability of various datasets in Australia, there is a notable gap in the existing research regarding the importance of Local Government Area (LGA) data in traffic incident duration prediction. Most studies have focused on national or state-level data, neglecting the localized characteristics of traffic incidents. This lack of granularity can lead to inaccurate predictions and ineffective incident response strategies.

To address this gap, this study aims to investigate the role of LGA data in traffic incident duration prediction using a comprehensive dataset of road traffic incidents in the Sydney Metropolitan Area. The dataset, described in "Data in Brief 51 (2023)" [1], offers a unique opportunity to explore the relationship between incident duration and LGA characteristics. By leveraging this dataset, we can provide insights into the factors that influence traffic incident duration and develop more accurate predictive models for traffic management authorities.

In this study, we undertake a comprehensive examination of the factors influencing the duration of traffic incidents within the Sydney Metropolitan Area. Our primary objective is to predict the duration of traffic incidents and classify them as either short-term or long-term. Our analyses reveal that models such as XGBoost and LightGBM notably outperform traditional approaches, with XGBoost achieving the lowest RMSE for predicting incident durations and the highest F1 score for classification tasks. We also conduct an extensive feature importance analysis using methods like tree split counts and SHAP values, highlighting the significance of factors such as the number of affected lanes, traffic volume, and types of vehicles involved.

By exploring the relationship between LGA data and traffic incident duration, this study aims to provide insights into the factors that influence traffic incident duration and develop more accurate predictive models for traffic management authorities.

The remainder of this paper is structured as follows: Section 2 reviews related works, providing a context for our research and highlighting existing studies in the domain. Section 3 delves into the case study, focusing on incident data and metadata from the Sydney metropolitan area. Section 4 explores the data, uncovering patterns and insights crucial for our analysis. Section 5 describes the methodology for our prediction models, detailing the approaches and techniques employed. Section 6 presents the results, showcasing the performance and accuracy of our models. In Section 7, we conduct a feature importance analysis to identify key factors influencing our predictions. Finally, Section 8 concludes the paper, summarising our findings and discussing potential future directions.



\section{Related Works}

Predicting the duration of incidents is a pivotal task for traffic incident management and it has garnered considerable attention in the literature. The predictions are often quantified in minutes or alternatively classified into categories such as short-term or long-term incidents. The demarcation between short and long-term incidents is a subject of much debate and hinges on a variety of factors, including the granularity and type of data available, the specific objectives of the study, and the geographic characteristics of the study area \cite{grigorev2022incident}. 

Incident duration prediction encompasses several critical timelines: the time required for emergency services or incident authorities to arrive at the scene, the time taken to clear the incident from the roadway, and the time until traffic flow returns to pre-incident conditions. Each of these aspects poses unique challenges and requires specific methodologies for accurate prediction. Notably, the prediction of traffic normalization post-incident demands an intricate understanding of the prevailing traffic conditions and the incident's broader impact on the traffic network. 
Furthermore, the literature reveals a trend towards the integration of advanced machine learning and data analytics techniques to enhance the accuracy and reliability of duration predictions. Studies increasingly leverage high-dimensional datasets, incorporating real-time traffic data, social media feeds, weather conditions, and other socio-economic factors to build comprehensive predictive models. This evolution in methodology reflects a growing recognition of the multifaceted nature of traffic incidents and the myriad factors that influence their duration. By adopting these advanced techniques, researchers aim to provide traffic management authorities with more precise and actionable predictions, thereby enabling more effective incident response strategies and mitigation measures.

In the last two decades, various approaches have been developed, involving different statistical models, machine learning approaches and data sources. 

Ozbay and Kachroo \cite{ozbay1999incident}, for example, applied decision trees to predict incident clearance times using independent variables like road hazards, property damage, personal injuries, broken trucks, vehicle fires, weather, etc. They used real data for testing and found that 60\% of the incidents had a prediction error of 10\% or less. However, Smith and Smith \cite{smith2002forecasting} found the stochastic model unsuitable due to the poor fit of Weibull and lognormal distributions. They achieved better results using non-parametric regression, and classification tree models, emphasizing the significance of tow truck response. 

In contrast, efforts by Chang and Chang \cite{Freeway2001Fuzzy} to use a classification tree model were less successful. While the classification accuracy for short incident durations was very high, the model struggled to similarly classify medium and long duration incidents. Recent works on traffic incident duration prediction utilise more advanced tree-based models \cite{OBAID2024107845, Shafiei2022_Data_and_simulation_Incidents, Grigorev2024_IEEETrans}.

In direct competition between different models, Valenti et al. \cite{valenti2010comparative} found that linear regression was the best approach for short duration incidents, with Relevance Vector Machine (RVM) models achieving the best prediction in the case of medium and medium to long duration incidents. Similarly, Yu et al. \cite{yubin} observed that Support Vector Machine (SVM) models outperformed Artificial Neural Networks (ANNs) for medium duration incidents.

Most of the existing literature on traffic incident duration predictions overlooks the role of location-specific traffic characteristics (like proximity of various amenities), with only 20\% of reviewed studies using data on traffic condition \cite{li2018overview}. However, majority of these studies have not yet achieved sufficient accuracy levels to offer real operational benefits, which underlines the role of statistical analysis. Feature importance estimation methods across traffic incident duration prediction studies include the use of XGBoost tree split analysis \cite{JIANG2021106431}, combination of tree-based prediction models with SHAP \cite{yang2021application, wang2024contributing}. Availability of data on traffic flow can also support or even enhance the incident duration prediction accuracy \cite{Shafiei2022_Data_and_simulation_Incidents, Grigorev2024_IEEETrans}.

Despite the use of detailed incident report datasets, majority of studies do not combine multiple types of data sources (i.e., social and economic data, event data, points-of-interest data, weather data, etc) \cite{Chapter5Usingmachinelearninganddeeplearningfortrafficcongestionpredictionareview} and rely primarily on incident report data. The combination of multiple types of features allows to perform more robust feature importance estimation, where incident-specific and indirectly related variables are compared against each other.

Considering the varied and complex nature of traffic incidents, several studies concentrate their research on certain types of incidents, such as vehicle collisions or mechanical failures and different aspects of incident timeline (clearance time, response time, total time of the incident) \cite{li2018overview}. This focused methodology permits an in-depth analysis of the distinct attributes inherent to these categories of incidents or duration intervals, recognizing that a universal approach might not effectively grasp the aspects of different incident categories. The rationale behind specializing in particular incident types is to increase the predictive accuracy of the models and enhance their effectiveness in situations where general models may not perform as well. On the contrary, the scarcity of detailed data on specific types of incidents can result in low prediction accuracy making the general analysis preferable.

\section{Case Study: Incident data and Metadata of Sydney Metropolitan Area}

For current study, we use the dataset described in "Data in Brief 51 (2023)" \cite{maparticle} offers a comprehensive overview of road traffic incidents (RTIs) within the Sydney Greater Metropolitan Area (GMA) in Australia, covering a 5.5-year period from 1 January 2017 to 31 July 2022. This public dataset is unique in its inclusion of specific data on the duration of each incident (RTI duration) and statistical data on local governmental areas where incidents occurred, capturing not just traffic crashes but also vehicle breakdowns and other events affecting traffic flow and safety. The dataset focuses on the Sydney Greater Metropolitan Area, New South Wales, Australia, divided into 333 zones based on Statistical Area Level-2 (SA2) classifications. The RTI data originates from the Open Data Hub, Transport for New South Wales (TfNSW), Australia. Supplemental data on road network metrics, bus network details, land use characteristics, and socioeconomic and travel characteristics were obtained from OpenStreetMap (using OSMnx and Rapidex), the Open Data Hub GTFS for NSW, and the Australian Bureau of Statistics, respectively.

The mean duration of incidents was approximately 39.65 minutes (see Table \ref{tab:stats}), suggesting a moderate average response time across all incidents. However, the standard deviation was high at 37.0 minutes, indicating substantial variability in incident durations. The shortest incident lasting only 0.62 minutes and the longest extending up to 251.46 minutes. The 25th percentile was 15.20 minutes, indicating that a quarter of the incidents were resolved in 15 minutes or less. The median duration was 28.68 minutes, and the 75th percentile was 50.88 minutes, showing that half of the incidents were resolved in approximately half an hour, and three-quarters were resolved within an hour.

\begin{table}[H]
\centering
\caption{Statistical Summary of Incident Duration within a Sydney Metropolitan Area}
\begin{tabular}{|c|c|}
\hline
\textbf{Statistic} & \textbf{Value} \\
\hline
Incident Count & 81142 \\
\hline
Mean & 39.65 \\
\hline
Standard Deviation & 37.0 \\
\hline
Minimum & 0.62 \\
\hline
25\% Quantile & 15.20 \\
\hline
Median (50\%) & 28.68 \\
\hline
75\% Quantile & 50.88 \\
\hline
Maximum & 251.46 \\
\hline
\end{tabular}
\label{tab:stats}
\end{table}

\begin{figure}[h]
    \centering
    \includegraphics[width=0.7\textwidth]{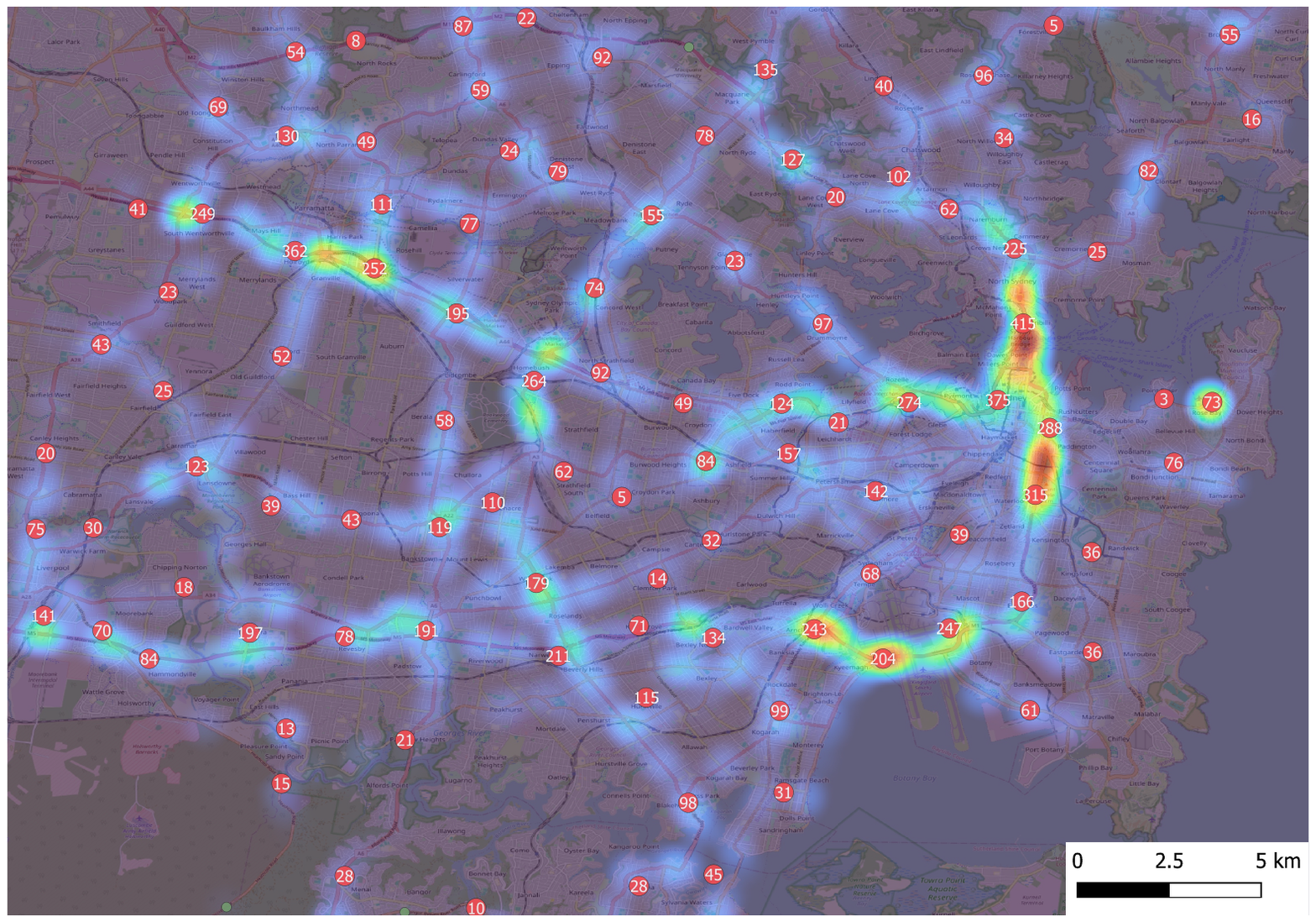}
    \caption{Heatmap of traffic incidents with marked centroids. Source: OpenStreetMap}
    \label{fig:heatmap}
\end{figure}

In Figure \ref{fig:heatmap}, the incident heat map shows the aggregate frequency of traffic incidents across different areas. Notably, a heightened concentration of incidents is visible at intersections. Furthermore, distinct areas exhibit exceptionally high frequencies of incidents, specifically the Harbour Bridge and the M1 motorway, including its associated roundabouts. 

\begin{figure}[h]
    \centering
    \includegraphics[width=0.9\textwidth]{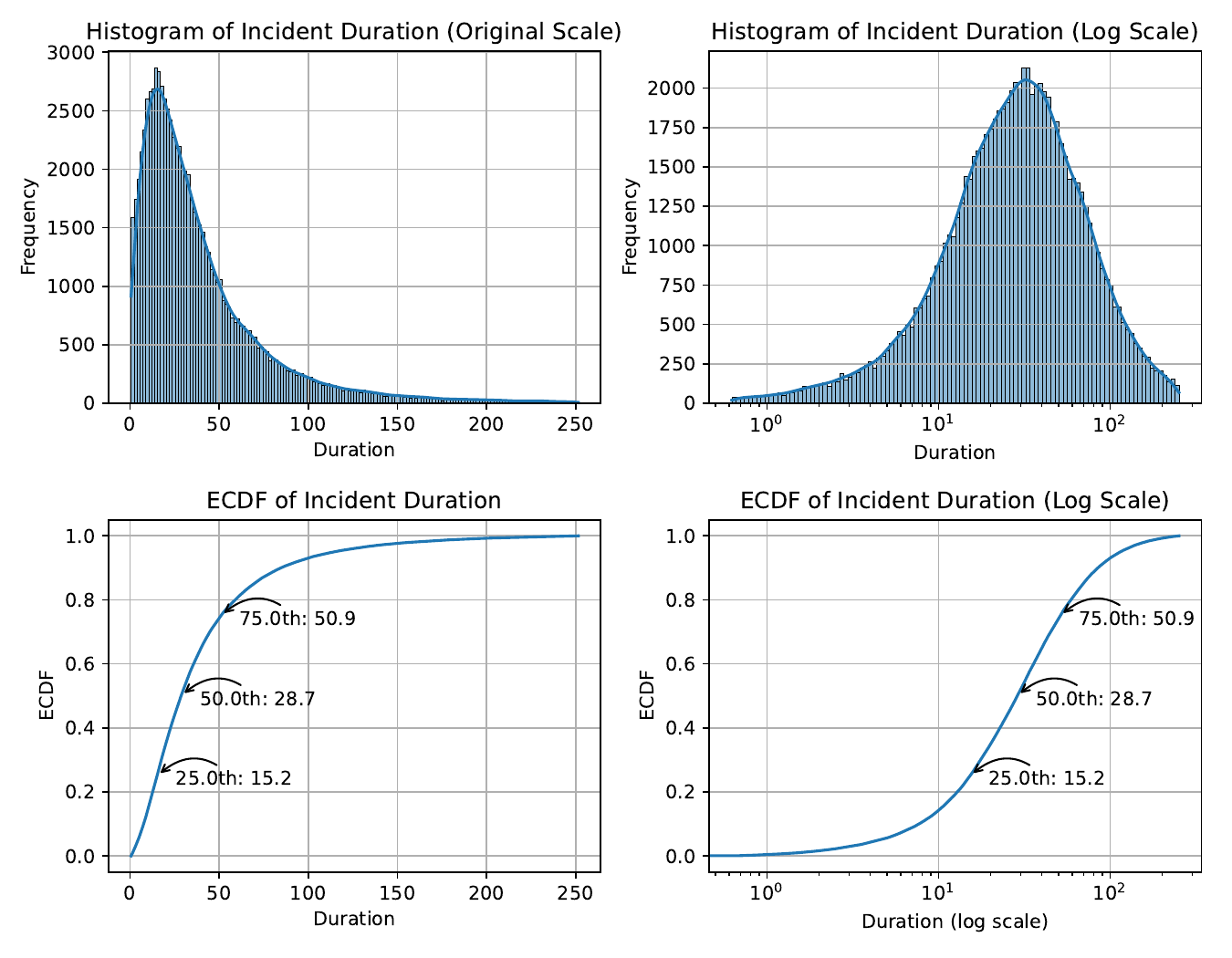}
    \caption{Incident duration statistics}
    \label{fig:durationplot}
\end{figure}

We observe that the incident duration is long-tail distributed (see Figure \ref{fig:durationplot}), which
is likely to affect the modelling accuracy due to the presence of extreme values. Various studies of traffic incident duration found the duration of incidents to follow log-normal or log-logistic distributions \cite{hojati2013hazard,Hazard2000}. For example, the use of logarithmic transformation can help normalize the data.

This RTI dataset (see Table \ref{tab:data_elements}) includes details on the location, start and end times, date, vehicles involved, type, and severity of the incidents, as well as information concerning the attending groups, advisories imposed, traffic status, and more. The road network characteristics encompass road type, road length, number of lanes, number of nodes and edges, road entropy, and node density, among others. Additional compiled secondary data include zonal boundaries, socio-economic attributes, and travel characteristics corresponding to Statistical Area Level-2 (SA2) geographic units. It facilitates a range of analyses, from patterns of crashes and breakdowns throughout the COVID-19 pandemic to the resilience of road networks against such incidents.

\begin{table}[h]
\scriptsize
    \centering
    \caption{Data Elements Description}
    \label{tab:data_elements}
    \begin{tabular}{|p{2cm}|p{6cm}|p{6cm}|}
    \hline
    \textbf{Variable} & \textbf{Description} & \textbf{Value Range/Samples} \\
    \hline
    1. Incident ID & Unique Incident id of each incident & 1001, 1002, 1003, ... \\
    \hline
    2. Main Category & Type of traffic incident: Breakdown, Crash, Others & "Breakdown", "Crash", "Others" \\
    \hline
    3. Longitude & GPS location of the incident in terms of longitude & -122.332, -71.087, ... \\
    \hline
    4. Latitude & GPS location of the incident in terms of latitude & 37.7749, 42.3601, ... \\
    \hline
    5. Start Time & Start date and time of the incidents & 01:01:2023 08:00, 02:02:2023 15:30, ... \\
    \hline
    6. Last Updated & Date and time of Last update about the Incident based on the issued advice & 01:01:2023 09:30, 02:02:2023 16:45, ... \\
    \hline
    7. Day & Which day of the week: Monday, Tuesday, Wednesday, Thursday, Friday, Saturday, Sunday & "Monday", "Tuesday", "Wednesday", ... \\
    \hline
    8. End Time & End date and time of the incident & 01:01:2023 08:30, 02:02:2023 16:00, ... \\
    \hline
    9. Duration in Minutes & Total duration of the incident & 30, 45, 60, ... \\
    \hline
    10. Primary Vehicle Category & Number of vehicles and type of vehicle involved in an incident & "Car", "Bus", "Truck", ... \\
    \hline
    11. Secondary Vehicle Category & Number of vehicles and type of vehicle involved in an incident & "Car", "Bus", "Truck", ... \\
    \hline
    12. Attending Groups & Type of attending group reached the incident location for clearing the traffic & "Police", "Emergency Services", ... \\
    \hline
    13. Display Name & Type of incident & "Accident", "Breakdown", ... \\
    \hline
    14. Is Major Incident & Binary variable indicating if the incident is major (1) or not (0) & "1" (Major Incident), "0" (Not Major) \\
    \hline
    15. Diversions & On which road or street, the traffic or approaching driver had been diverted? & "Main Street", "Highway A", ... \\
    \hline
    16. Advice A & Advisory message displayed to road users due to the incident & "Expect delays", "Use alternative route", ... \\
    \hline
    17. Advice B & Another advisory message displayed to road users after certain actions by authorities & "Lane closure in effect", "Follow detour signs", ... \\
    \hline
    18. Other Advice & Additional advisory messages displayed to road users after certain actions by authorities & "Emergency services on site", "Road cleared", ... \\
    \hline
    19. Closure Type & Effect on the lane(s): Affected, Lanes closed/Closed, Unknown & "Affected", "Lanes closed", "Unknown" \\
    \hline
    20. Direction & Direction of travel of the vehicle involved in an incident & "Eastbound", "Westbound", "Inbound", ... \\
    \hline
    21. Main Street & Street information & "Main Street", "Highway A", ... \\
    \hline
    22. Affected Lane & Number of lanes affected due to the incident & 1, 2, 3, ... \\
    \hline
    23. Actual Number of Lanes & Number of lanes of the street on which the incident has occurred & 2, 4, 6, ... \\
    \hline
    24. Suburb & Suburb name & "Downtown", "Uptown", ... \\
    \hline
    25. Traffic Volume & Existing traffic volume at the time of the incident: Heavy, Light, Moderate, Unknown & "Heavy", "Light", "Moderate", "Unknown" \\
    \hline
    26. SA2 CODE21 & Statistical Area Level-2, Zone ID & 1001, 1002, 1003, ... \\
    \hline
    27. SA2 NAME21 & Statistical Area Level-2, Zone name & "Central Business District", "Suburb A", ... \\
    \hline
    28. SA3 CODE21 & Statistical Area Level-3, Zone ID & 2001, 2002, 2003, ... \\
    \hline
    29. SA3 NAME21 & Statistical Area Level-3, Zone name & "Metropolitan Area A", "Metropolitan Area B", ... \\
    \hline
    30. SA4 CODE21 & Statistical Area Level-4, Zone ID & 3001, 3002, 3003, ... \\
    \hline
    31. SA4 NAME21 & Statistical Area Level-4, Zone name & "City Region A", "City Region B", ... \\
    \hline
    32. AREASQKM21 & Area of each zone & 50.2, 75.6, 100.0, ... \\
    \hline
    33. LOCI URI21 & Url links for the statistical area level-2 & "https://example.com/zone-1", "https://example.com/zone-2", ... \\
    \hline
    \end{tabular}
\end{table}

The metadata (see Table \ref{tab:metadata_sydney}) includes a wide range of urban and socio-economic indicators:

1. Urban Structure and Land Use: Entries capture various measures related to urban planning and land use, including total area (Sqkm), lengths of different types of roads (Motorway, Trunk, Primary, Secondary, Tertiary, Residential, Living Street, and Unclassified Roads), total road length, and specific area types (Industrial, Parkland, Primary Production, Residential, Transport, Water body Areas).

2. Transportation and Mobility: This includes the entropy of road networks, the total number of nodes (intersections), types of nodes (dead ends, and nodes connected to 2, 3, or 4 links), and metrics related to public and private transportation modes. It covers indicators such as the number of bus stops, and the number of people traveling to work by various means (public transit, taxi/rideshare, car—either as a driver or passenger, and other methods).

3. Socio-Economic Indicators: The table provides detailed socio-economic data such as the total population, percentages of dwellings based on the number of motor vehicles, unemployment percentages, average monthly income, work-from-home statistics, and job sector distribution (white-collar vs. blue-collar jobs). It also includes family-oriented data like the average number of persons in a family household.

\begin{table}[h]
    \scriptsize
    \centering
    \caption{Metadata for Greater Sydney Metropolitan Area}
    \label{tab:metadata_sydney}
    \begin{tabular}{|p{1.0cm}|p{1.5cm}|p{4cm}|p{1.0cm}|p{1.5cm}|p{4cm}|}
    \hline
    \textbf{Index} & \textbf{Short Name} & \textbf{Long Name} & \textbf{Index} & \textbf{Short Name} & \textbf{Long Name} \\
    \hline
    \specialrule{.1em}{.05em}{.05em}
    0 & ZID & Zone ID of Statistical Area Level-2 (SA2\_CODE21) & 25 & IA & Industrial Area (km2) \\
    \specialrule{.05em}{.05em}{.05em}
    1 & Area & Area in Sqkm & 26 & OA & Other Area (km2) \\
    \specialrule{.05em}{.05em}{.05em}
    2 & ML & Motorway Length (km) & 27 & PA & Parkland Area (km2) \\
    \specialrule{.05em}{.05em}{.05em}
    3 & TRL & Trunk Road Length (km) & 28 & PrA & Primary Production Area (km2) \\
    \specialrule{.05em}{.05em}{.05em}
    4 & PRL & Primary Road Length (km) & 29 & RA & Residential Area (km2) \\
    \specialrule{.05em}{.05em}{.05em}
    5 & SRL & Secondary Road Length (km) & 30 & TA & Transport Area (km2) \\
    \specialrule{.05em}{.05em}{.05em}
    6 & TrRL & Tertiary Road Length (km) & 31 & WbA & Water body Area (km2) \\
    \specialrule{.05em}{.05em}{.05em}
    7 & RRL & Residential Road Length (km) & 32 & EoLU & Entropy of Land Use \\
    \specialrule{.05em}{.05em}{.05em}
    8 & LsRL & Living Street Road Length (km) & 33 & TP & Total Population \\
    \specialrule{.05em}{.05em}{.05em}
    9 & URL & Unclassified Road Length (km) & 34 & PD0MV & Percentage of Dwellings with No Motor Vehicles \\
    \specialrule{.05em}{.05em}{.05em}
    10 & ToRL & Total Road Length (km) & 35 & PD1MV & Percentage of Dwellings with One Motor Vehicle \\
    \specialrule{.05em}{.05em}{.05em}
    11 & EoR & Entropy of road & 36 & PD2MV & Percentage of Dwellings with Two Motor Vehicles \\
    \specialrule{.05em}{.05em}{.05em}
    12 & NoN & Number of Nodes & 37 & PD3MV & Percentage of Dwellings with Three Motor Vehicles \\
    \specialrule{.05em}{.05em}{.05em}
    13 & NDEs & Number of Dead Ends/Number of Cul-de-sacs & 38 & PD4MV & Percentage of Dwellings with Four or More Motor Vehicles \\
    \specialrule{.05em}{.05em}{.05em}
    14 & NNC2L & Number of Nodes Connected to 2 Links & 39 & PUE & Percentage of Unemployment \\
    \specialrule{.05em}{.05em}{.05em}
    15 & NNC3L & Number of Nodes Connected to 3 Links & 40 & AMI & Average Monthly Income in AUD \\
    \specialrule{.05em}{.05em}{.05em}
    16 & NNC4L & Number of Nodes Connected to 4 Links & 41 & NPTbyPT & Number of people travel to work by Public Transit \\
    \specialrule{.05em}{.05em}{.05em}
    17 & AND & Average Node Degree & 42 & NPTtWbyTx & Number of People Travel to Work by One Method\_Taxi/Rideshare \\
    \specialrule{.05em}{.05em}{.05em}
    18 & NE & Number of Links/Edges & 43 & NPTtWbyCD & Number of People Travel to Work by One Method\_Car as Driver \\
    \specialrule{.05em}{.05em}{.05em}
    19 & MCI & Meshedness Coefficient Index & 44 & NPTtWbyCP & Number of People Travel to Work by One Method\_Car as Passenger \\
    \specialrule{.05em}{.05em}{.05em}
    20 & CoI & Completeness Index & 45 & NPTtWbyO & Number of People Travel to Work by One Method\_Other \\
    \specialrule{.05em}{.05em}{.05em}
    21 & NBS & Number of Bus Stops & 46 & NPWfH & Number of People Work from Home \\
    \specialrule{.05em}{.05em}{.05em}
    22 & CA & Commercial Area (km2) & 47 & PWCJH & Percentage of White Collar Job Holders \\
    \specialrule{.05em}{.05em}{.05em}
    23 & EA & Education Area (km2) & 48 & PBCJH & Percentage of Blue Collar Job Holders \\
    \specialrule{.05em}{.05em}{.05em}
    24 & HA & Hospital/Medical Area (km2) & 49 & ANP\_FH & Average Number Persons in a Family Household \\
    \specialrule{.1em}{.05em}{.05em}
    \end{tabular}
\end{table}

In total, the merged data set, where each incident data associated with corresponding socio-economic and road network data contains 82 variables for each of 85,612 incident records.


\section{Data exploration}

\begin{figure}[h]
    \centering
    \begin{subfigure}[b]{0.45\textwidth}
        \centering
        \includegraphics[width=\textwidth]{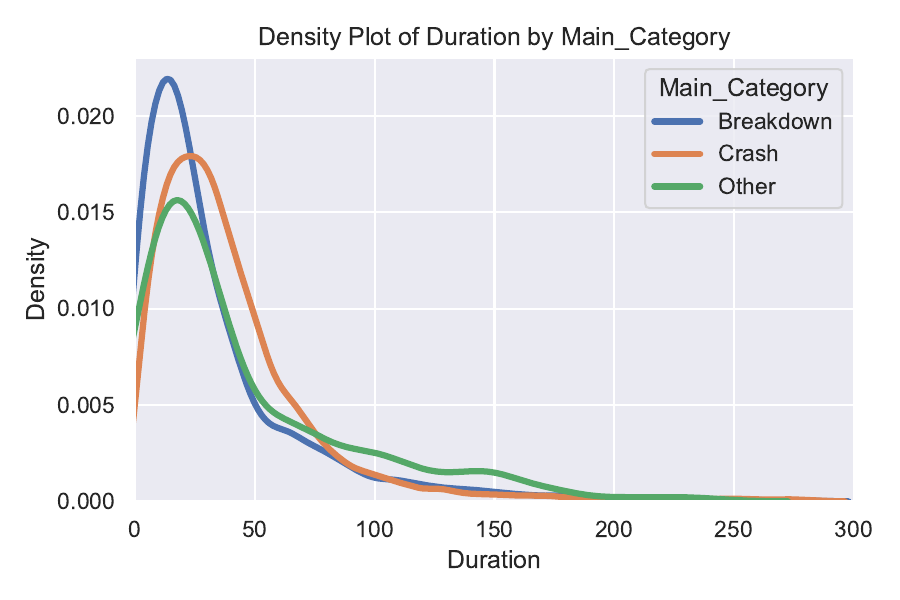}
        \caption{Duration by Main Category}
        \label{fig:density_Main_Category}
    \end{subfigure}
    \hfill
    \begin{subfigure}[b]{0.45\textwidth}
        \centering
        \includegraphics[width=\textwidth]{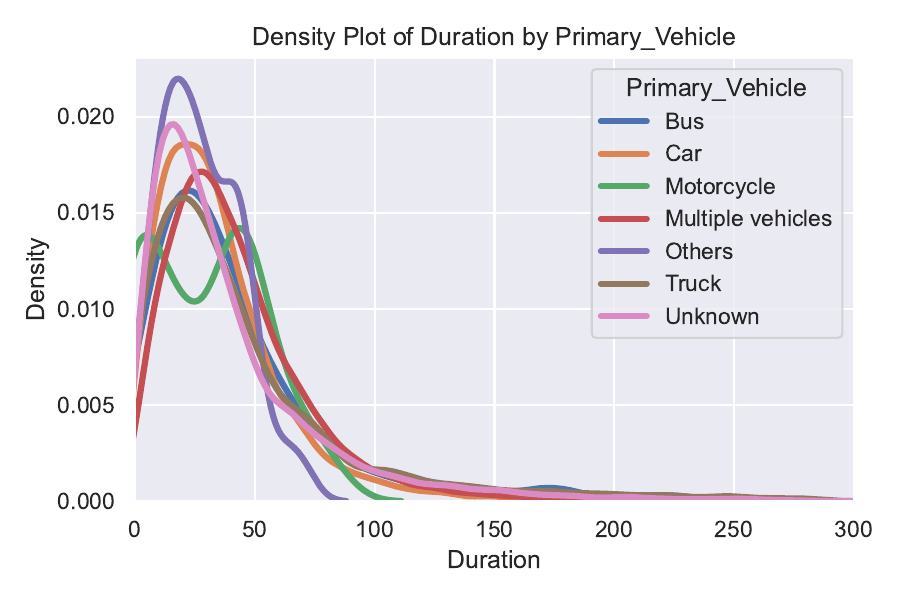}
        \caption{Duration by Primary Vehicle}
        \label{fig:density_Primary_Vehicle}
    \end{subfigure}
    \caption{Density Plots of Duration by Main Category and Primary Vehicle}
\end{figure}

The density plots of incident duration provide valuable insights into how various factors influence the time taken to resolve incidents. Figure \ref{fig:density_Main_Category} demonstrates the density plot of incident duration by Main Category, revealing that traffic crashes are noticeably skewed towards longer durations in comparison to breakdowns. Similarly, Figure \ref{fig:density_Primary_Vehicle} shows the duration pattern by Primary Vehicle, highlighting characteristic skewness for each vehicle type—for example, incidents involving motorcycles peak at shorter durations, whereas incidents with multiple vehicles skew towards longer durations.

\begin{figure}[h]
    \centering
    \begin{subfigure}[b]{0.45\textwidth}
        \centering
        \includegraphics[width=\textwidth]{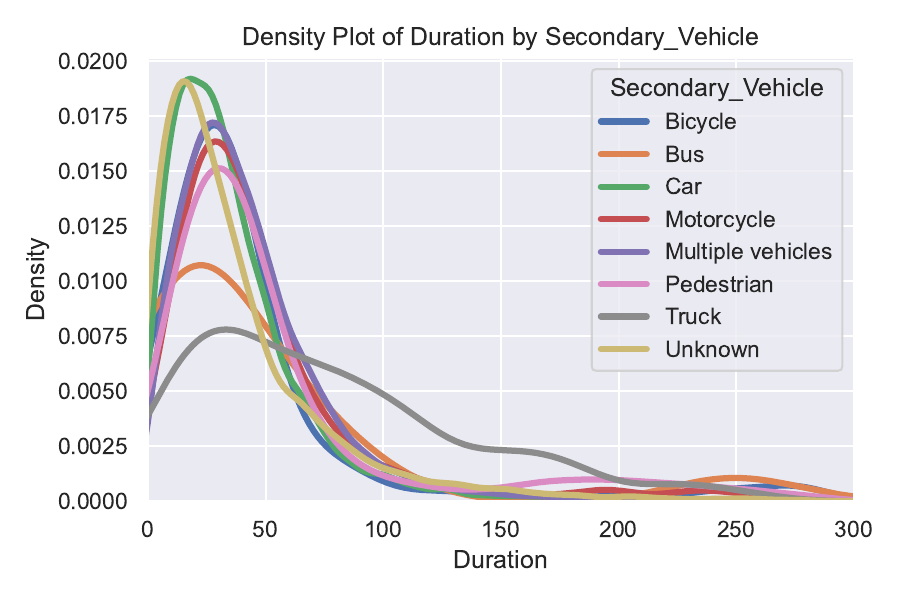}
        \caption{Duration by Secondary Vehicle}
        \label{fig:density_Secondary_Vehicle}
    \end{subfigure}
    \hfill
    \begin{subfigure}[b]{0.45\textwidth}
        \centering
        \includegraphics[width=\textwidth]{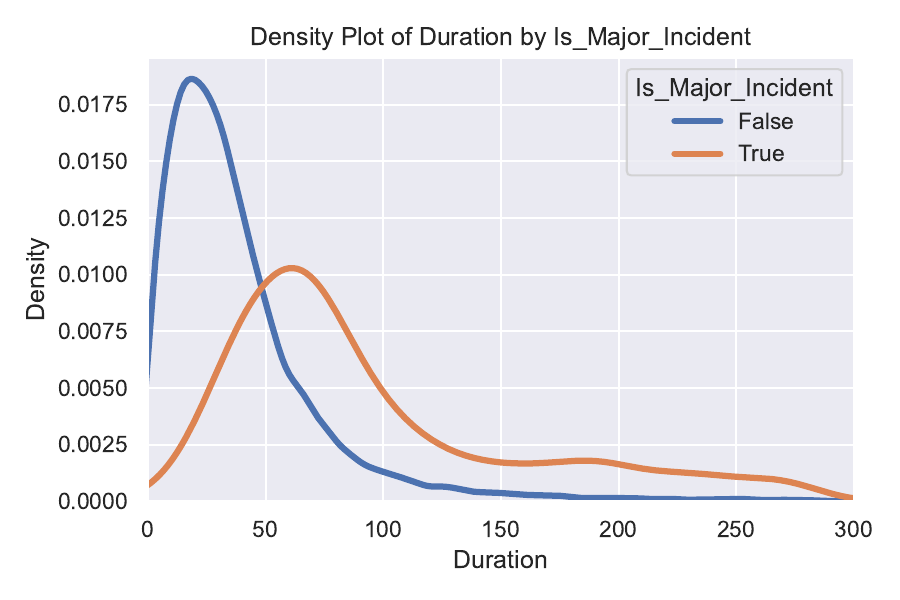}
        \caption{Duration by Is Major Incident}
        \label{fig:density_Is_Major_Incident}
    \end{subfigure}
    \caption{Density Plots of Duration by Secondary Vehicle and Is Major Incident}
\end{figure}

Figure \ref{fig:density_Secondary_Vehicle} underscores the significant increase in incident duration when a secondary vehicle truck is involved, whereas incidents with a car as the secondary vehicle tend to have shorter durations. The density plot of incident duration by incident scale, shown in Figure \ref{fig:density_Is_Major_Incident}, indicates that major incidents take significantly more time to resolve, with durations typically skewed towards the 100-300 minutes range.

\begin{figure}[h]
    \centering
    \begin{subfigure}[b]{0.45\textwidth}
        \centering
        \includegraphics[width=\textwidth]{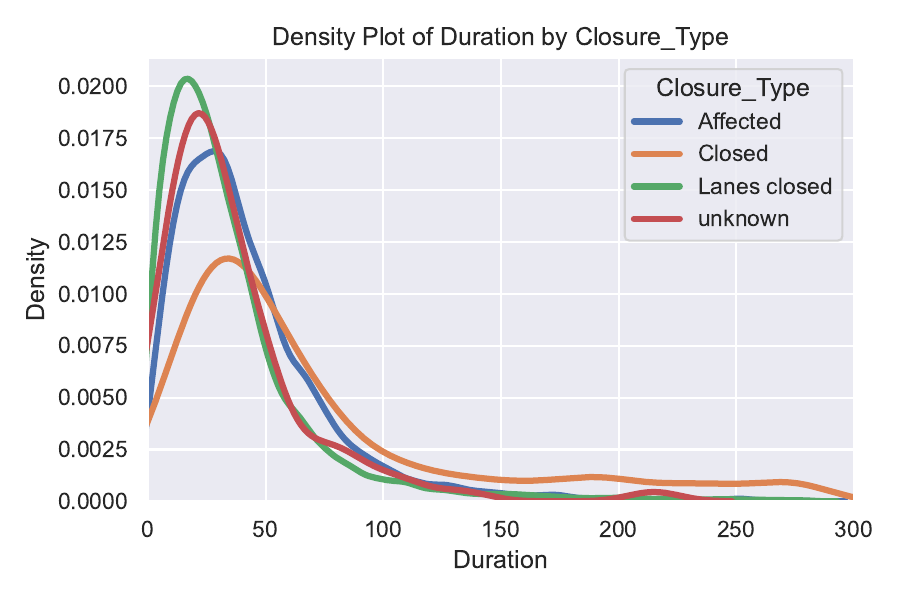}
        \caption{Duration by Closure Type}
        \label{fig:density_Closure_Type}
    \end{subfigure}
    \hfill
    \begin{subfigure}[b]{0.45\textwidth}
        \centering
        \includegraphics[width=\textwidth]{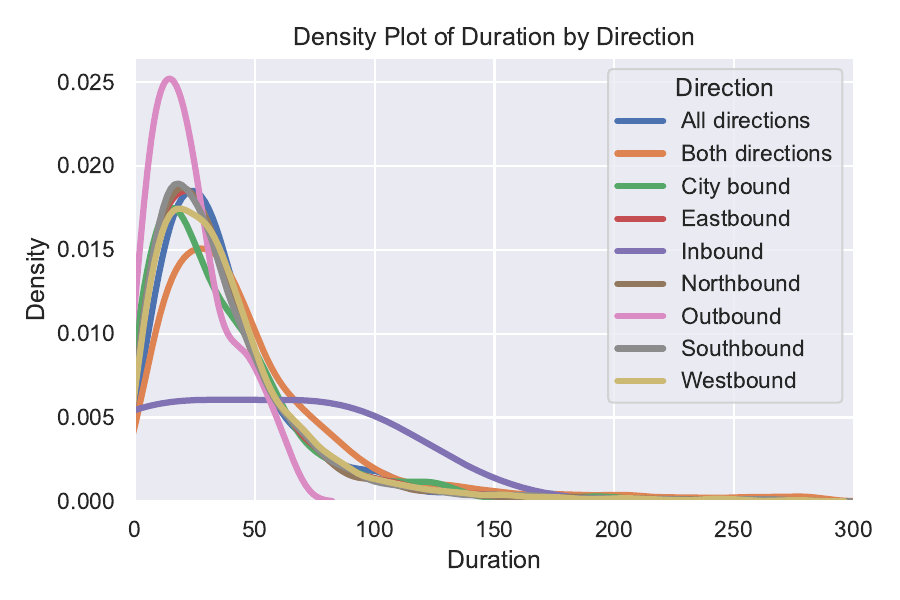}
        \caption{Duration by Direction}
        \label{fig:density_Direction}
    \end{subfigure}
    \caption{Density Plots of Duration by Closure Type and Direction}
\end{figure}

Figure \ref{fig:density_Closure_Type} illustrates that when a lane is closed, the incident duration generally extends longer. Directionality of incidents also plays a role, as depicted in Figure \ref{fig:density_Direction}; incidents in the inbound direction are associated with a substantial increase in duration.

\begin{figure}[h]
    \centering
    \begin{subfigure}[b]{0.6\textwidth}
        \centering
        \includegraphics[width=\textwidth]{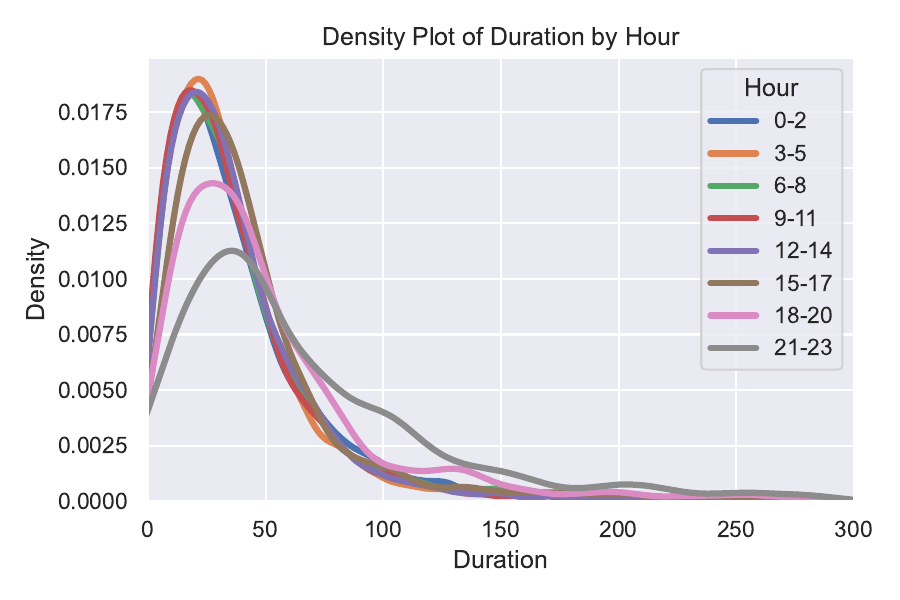}
        \caption{Duration by Hour}
        \label{fig:density_Hour}
    \end{subfigure}
    \caption{Density Plot of Duration by Hour}
\end{figure}

Lastly, Figure \ref{fig:density_Hour} categorizes incident durations by 3-hour intervals, revealing longer durations for incidents occurring during evening hours (after 6 pm). Other density plots, which considered factors such as traffic volume, the number of vehicles involved, and month, did not yield significant insights and indicated that incidents follow similar duration patterns across these additional dimensions.

\section*{Heatmaps of Wasserstein distances}

When exploring and comparing the distributions of incident durations across various categorical variables, an intuitive approach is to utilize Wasserstein distance estimation. The Wasserstein distance (also known as Earth Mover's Distance) is a measure of the distance between two probability distributions \cite{givens1984class, vallender1974calculation}. It provides a quantitative way to compare the shapes and positions of distributions, which is particularly useful in highlighting differences in distributions of incident durations across different categorical groups.

The Wasserstein distance arises from optimal transport theory, which is concerned with finding the most efficient way to reshape one probability distribution into another. This concept can be visualized as the minimum "work" needed to transport a pile of dirt configured in one shape to match another shape.

Mathematically, for two probability distributions \(\mu\) and \(\nu\) defined on a metric space \(M\) with a distance metric \(d\), the \(p\)-th Wasserstein distance (\(W_p\)) is defined as:

\[
W_p(\mu, \nu) = \left( \inf_{\gamma \in \Gamma(\mu, \nu)} \int_{M \times M} d(x, y)^p \, d\gamma(x, y) \right)^{1/p},
\]

where \(\Gamma(\mu, \nu)\) denotes the set of all joint distributions \(\gamma(x, y)\) whose marginals are \(\mu\) and \(\nu\), respectively. Here, \(d(x, y)\) represents the distance between points \(x\) and \(y\).

For practical purposes, the 1\textsuperscript{st} Wasserstein Distance (\(W_1\)), also known simply as the Wasserstein distance, is often used:

\[
W_1(\mu, \nu) = \inf_{\gamma \in \Gamma(\mu, \nu)} \int_{M \times M} d(x, y) \, d\gamma(x, y).
\]

In the context of incident duration analysis, we deal with empirical distributions of durations based on various categorical variables (e.g., Main Category, Primary Vehicle, Secondary Vehicle, etc.). To measure and compare the differences between these distributions, we compute the Wasserstein distances between the empirical distributions of incident durations for each pair of categorical values.

Let \(X_{\text{cat}_1}\) and \(X_{\text{cat}_2}\) represent the incident durations for two different categories of a variable. Let \(F_{\text{cat}_1}\) and \(F_{\text{cat}_2}\) denote their respective empirical cumulative distribution functions (ECDFs). The Wasserstein distance \(W_1\) between these two empirical distributions can be approximated as:

\[
W_1(F_{\text{cat}_1}, F_{\text{cat}_2}) = \int_{-\infty}^{\infty} \left| \text{ECDF}_{\text{cat}_1}(x) - \text{ECDF}_{\text{cat}_2}(x) \right| \, dx.
\]

This equation computes the area between the ECDFs of the two distributions, thus providing a measure of the discrepancy between incident durations across different categorical values.

\begin{figure}[h]
    \centering
    \includegraphics[width=0.7\textwidth]{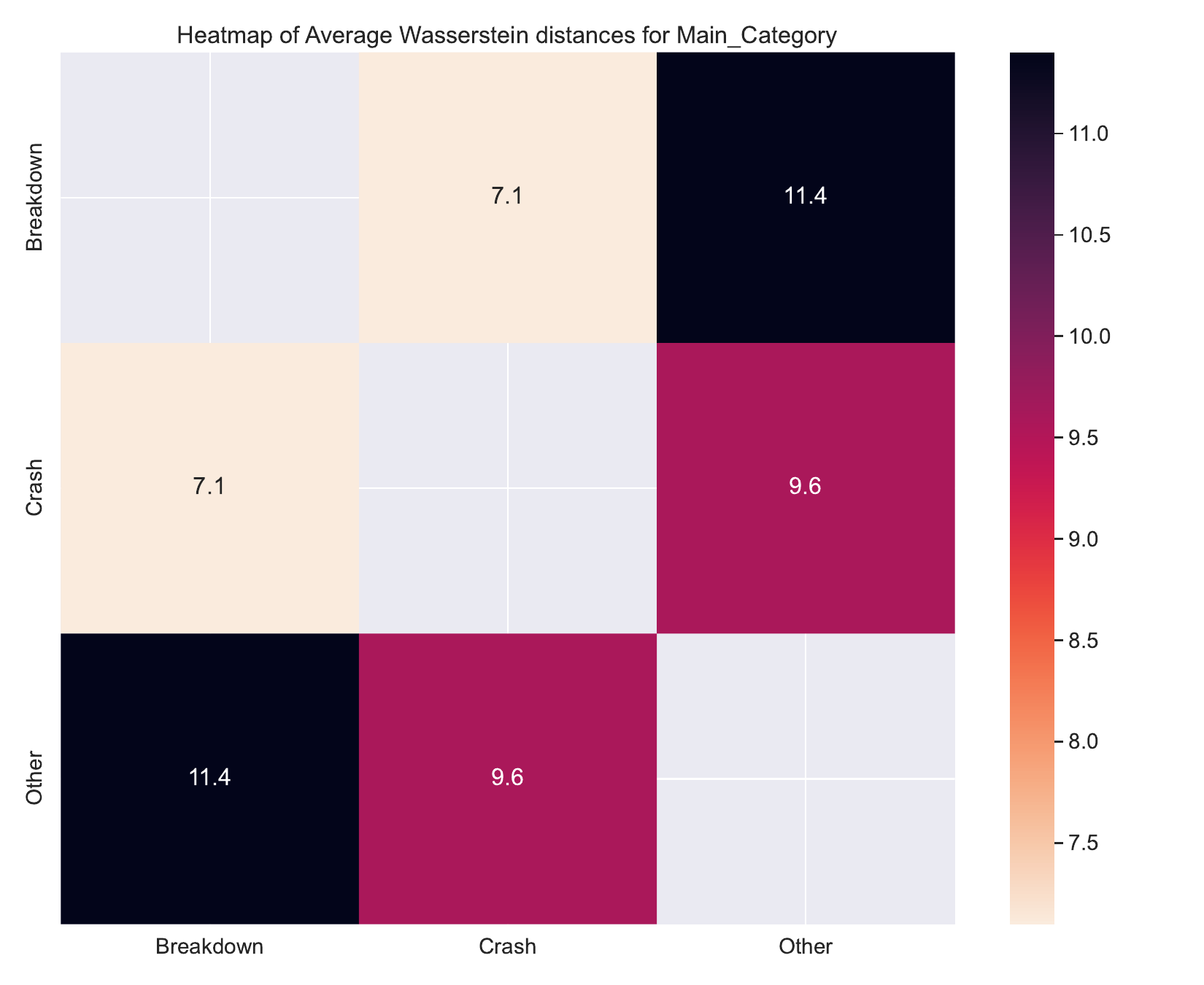}
    \caption{Heatmap of Average Wasserstein distances for Main Category}
    \label{fig:heatmap_Main_Category}
\end{figure}

Figure \ref{fig:heatmap_Main_Category} shows the heatmap of average Wasserstein distances for Main Category. The color denotes the degree of difference between incident duration distribution at two different values of the categorical variable. For example, duration distribution of breakdowns is slightly different from crashes and significanlty different from other types of incidents.

\begin{figure}[h]
    \centering
    \includegraphics[width=0.9\textwidth]{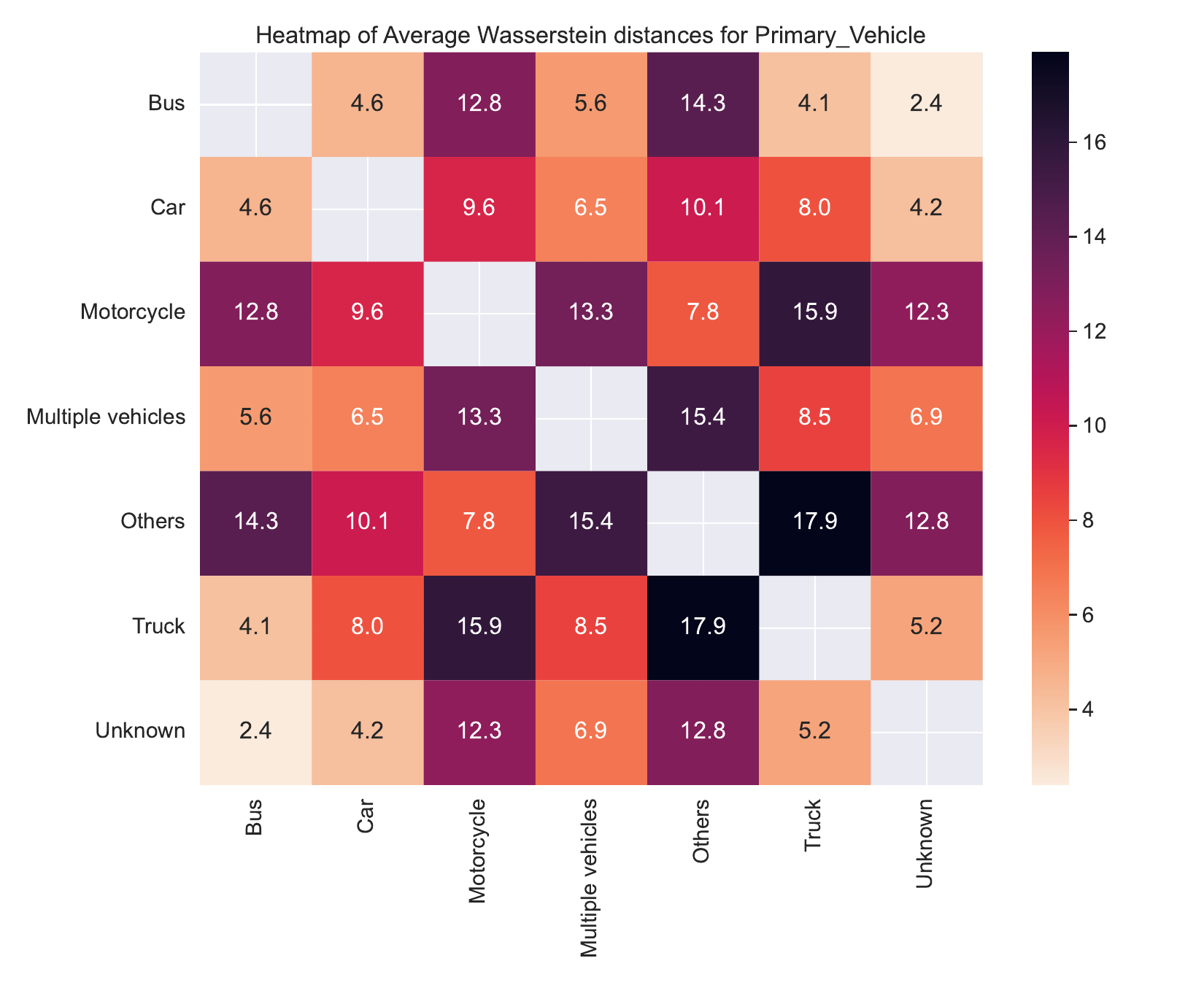}
    \caption{Heatmap of Average Wasserstein distances for Primary Vehicle}
    \label{fig:heatmap_Primary_Vehicle}
\end{figure}

Figure \ref{fig:heatmap_Primary_Vehicle} shows the heatmap of average Wasserstein distances for Primary Vehicle. Motorcycles and trucks show significant difference in incident duration distribution, while buses and trucks follow a similar pattern. Multiple vehicle crashes also have noticeable relative similarity with multi-vehciles crashes (WD=5.6). Unknown type of primary vehicle has very similar pattern to buses.

\begin{figure}[h]
    \centering
    \includegraphics[width=0.9\textwidth]{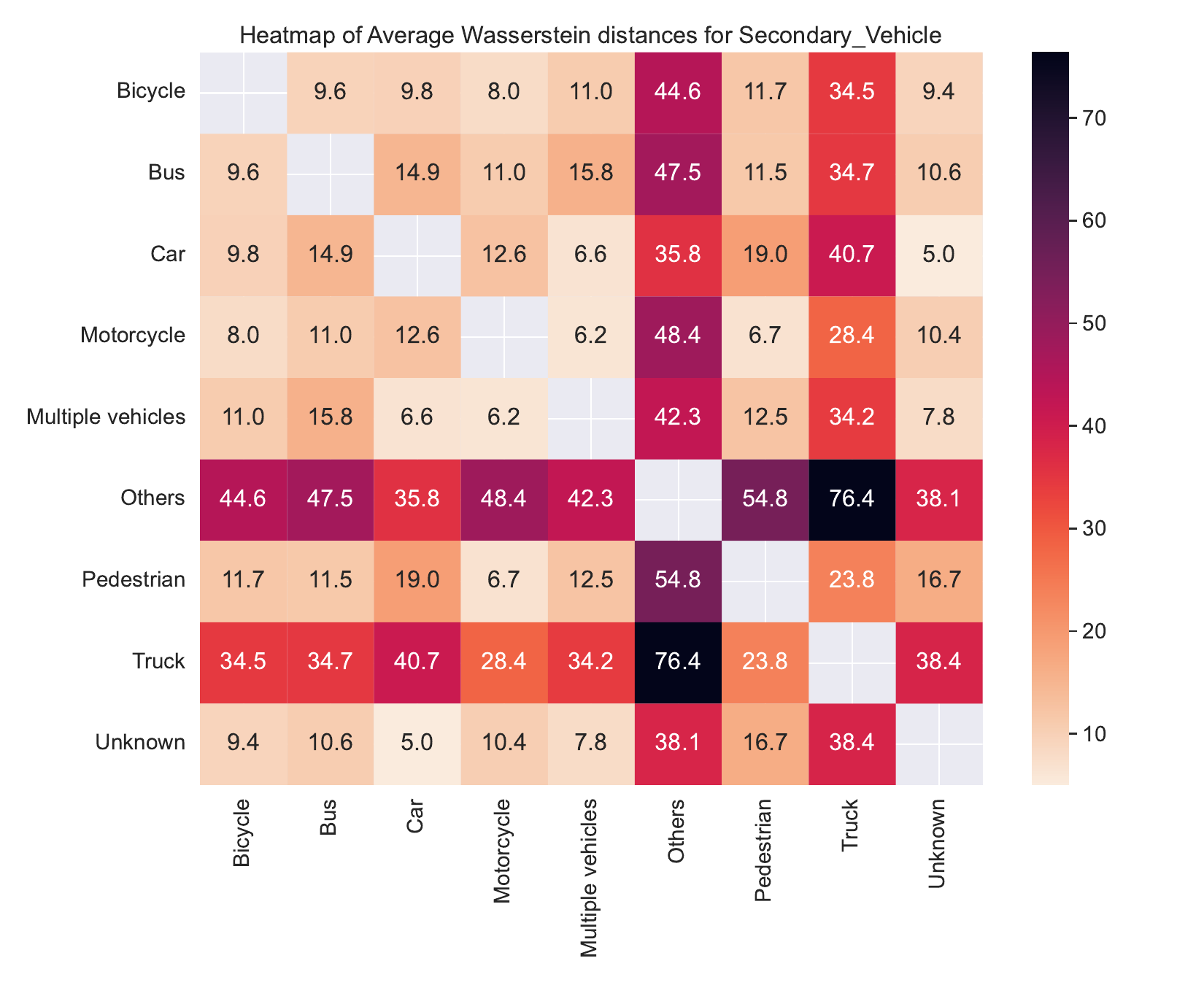}
    \caption{Heatmap of Average Wasserstein distances for Secondary Vehicle}
    \label{fig:heatmap_Secondary_Vehicle}
\end{figure}

Figure \ref{fig:heatmap_Secondary_Vehicle} shows the heatmap of average Wasserstein distances for Secondary Vehicle. While duration pattern between secondary vehicles is mostly similar, trucks show significant difference in duration in comparison to all other types of vehicles. Bicycle incidents are the most similar to motorcycle in duration pattern.

\begin{figure}[h]
    \centering
    \includegraphics[width=0.9\textwidth]{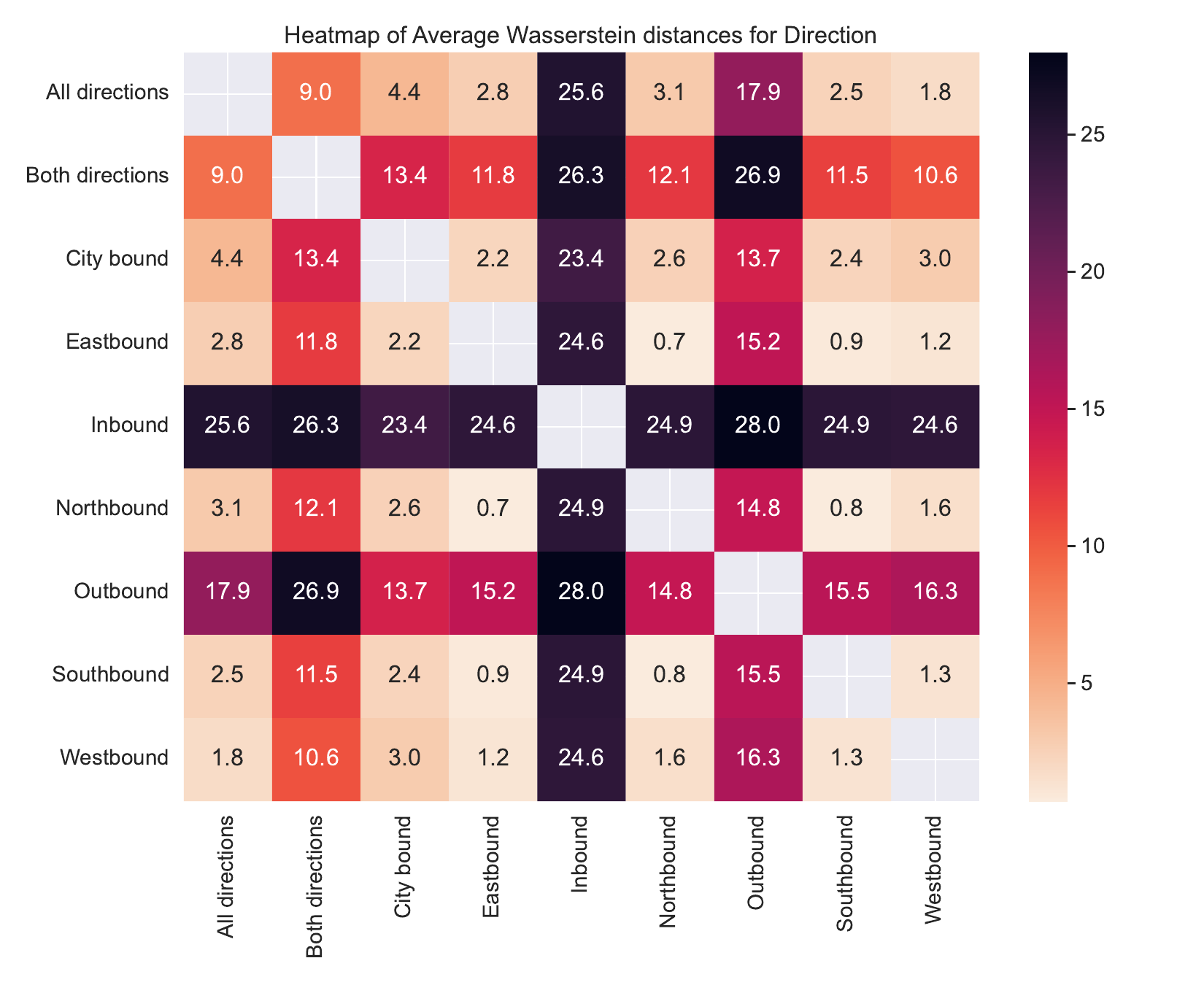}
    \caption{Heatmap of Average Wasserstein distances for Direction}
    \label{fig:heatmap_Direction}
\end{figure}

Figure \ref{fig:heatmap_Direction} shows the heatmap of average Wasserstein distances for Direction. The inbound traffic direction shows the highest difference from other types of directions according to duration pattern. Moderate difference is also observed for outbound traffic direction.

\begin{figure}[h]
    \centering
    \includegraphics[width=0.8\textwidth]{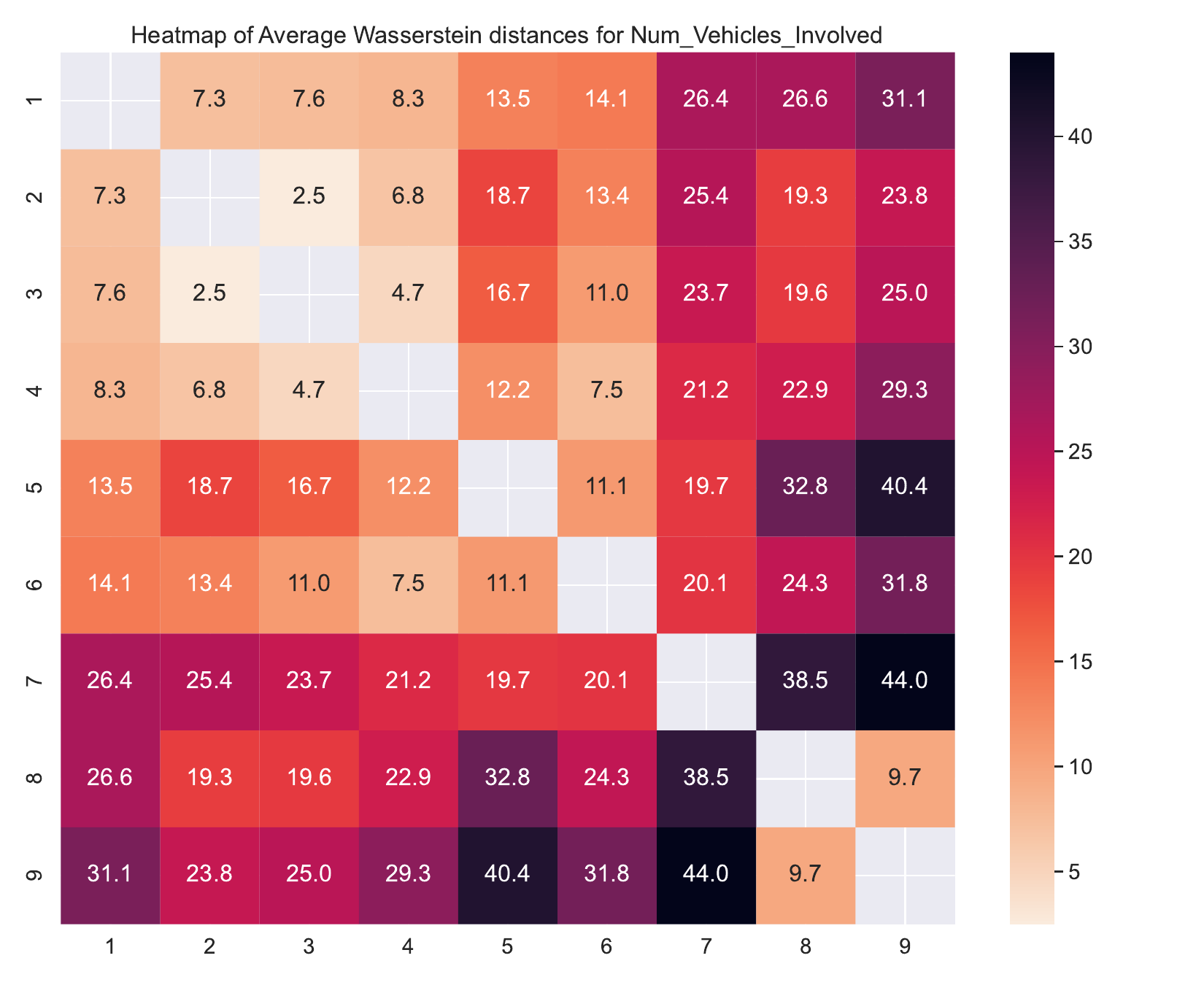}
    \caption{Heatmap of Average Wasserstein distances for Num Vehicles Involved}
    \label{fig:heatmap_Num_Vehicles_Involved}
\end{figure}

Figure \ref{fig:heatmap_Num_Vehicles_Involved} shows the heatmap of average Wasserstein distances for Number of Vehicles Involved. There is a low amount of difference observed for incident duration patterns across small numbers of vehicles involved. Significant deviation of patterns starts when at least 5 vehicles involved in incident, with extreme variation in duration patterns for incidents involving 7 to 9 vehicles.

\begin{figure}[h]
    \centering
    \includegraphics[width=0.8\textwidth]{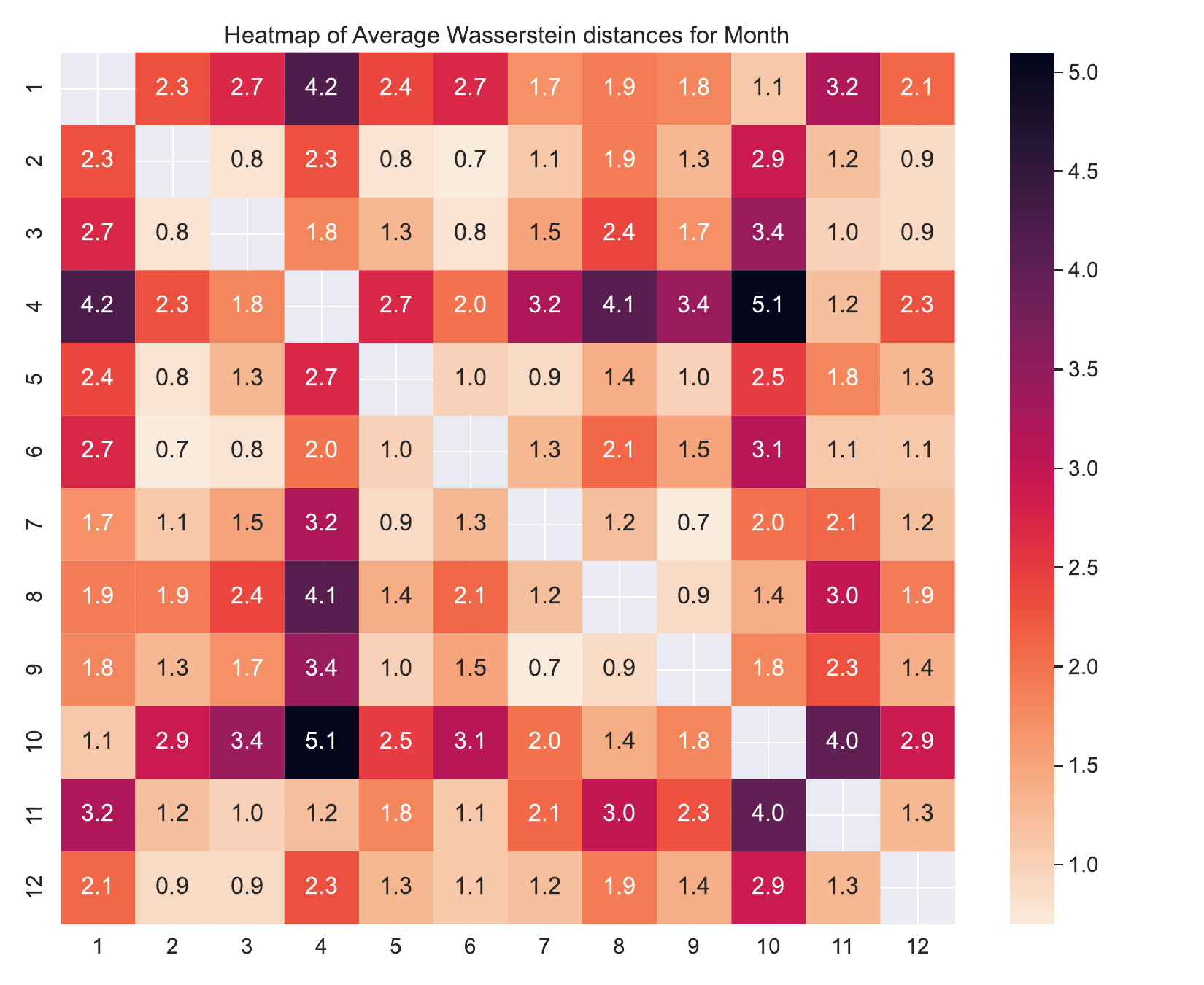}
    \caption{Heatmap of Average Wasserstein distances for Month}
    \label{fig:heatmap_Month}
\end{figure}

Figure \ref{fig:heatmap_Month} shows the heatmap of average Wasserstein distances for Month. For unknown reason, incidents occured during January, April and October have significantly different incident duration patterns from other months, with April having the highest degree of difference in total.

\begin{figure}[h]
    \centering
    \includegraphics[width=0.9\textwidth]{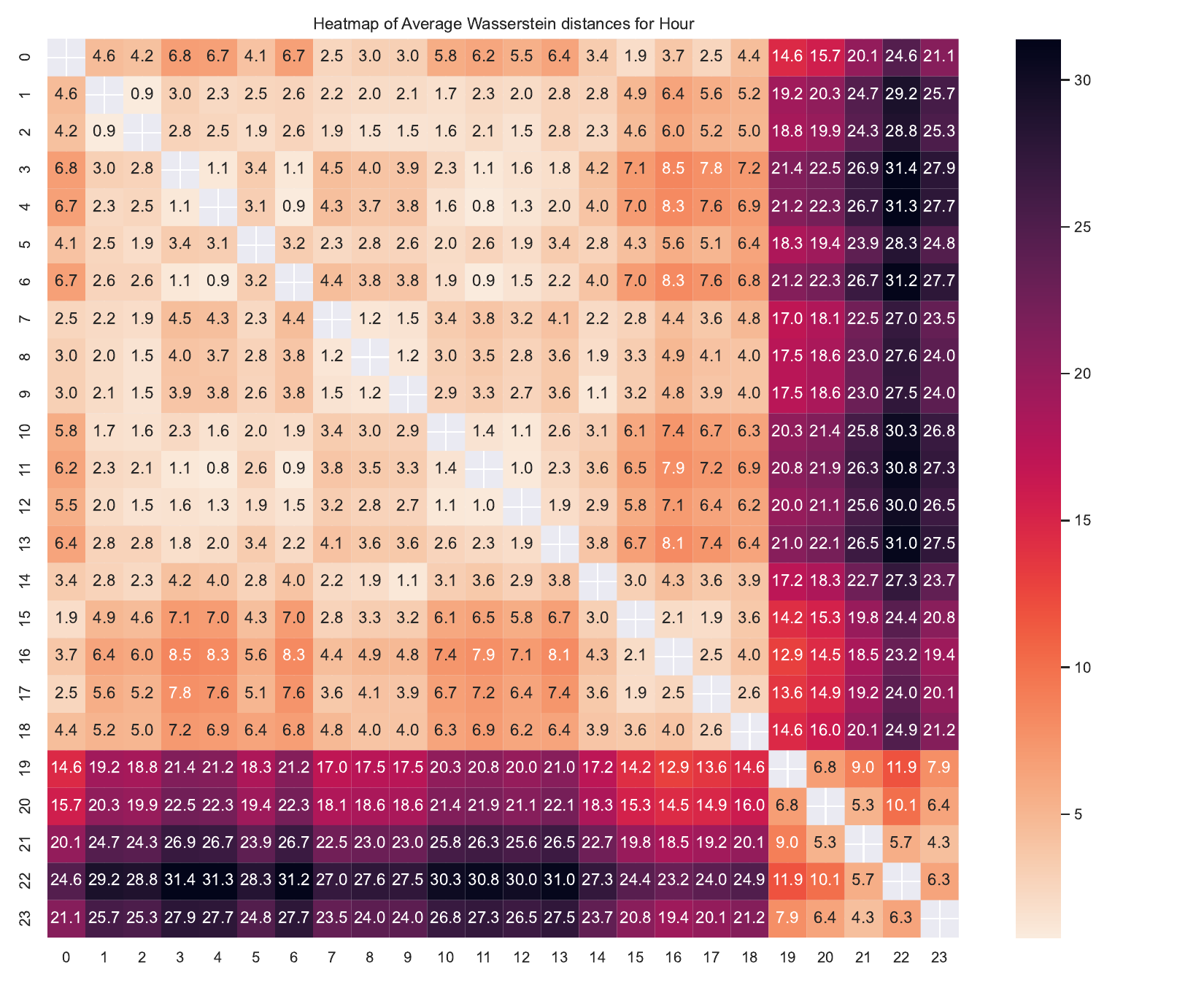}
    \caption{Heatmap of Average Wasserstein distances for Hour}
    \label{fig:heatmap_Hour}
\end{figure}

Figure \ref{fig:heatmap_Hour} shows the heatmap of average Wasserstein distances for Hour. Hours from 0 to 18 show no significant difference in duration pattern. Duration patterns of evening hours have significant difference from early hours but not between themselves. It may imply the change in incident resolution mode after 6 PM. Difference in duration patterns between after-hours and early hours drops at 7 to 9 AM and 3 to 5 PM.

The analysis of incident duration using density plots and Wasserstein distances provides substantial insights into the dynamics of incident resolution. By examining density plots, we observed that incidents' durations differ significantly depending on various factors such as the main category of the incident, type of primary and secondary vehicles involved, whether an incident is major, lane closure, direction, and hour of occurrence. These visualizations highlighted characteristic patterns, such as traffic crashes tending towards longer durations and incidents involving trucks or multiple vehicles also exhibiting extended durations. 

By using the Wasserstein distance metric, we could effectively demonstrate how empirical distributions of incident durations vary across different categorical variables. For example, the empirical distributions of incident durations highlighted greater discrepancies in the form of larger Wasserstein distances among certain categories like primary and secondary vehicle types, number of vehicles involved, and specific months and hours.

\section{Methodology}

In our approach we are predicting the duration of incidents and classifying them into short-term and long-term categories. We allocate 80\% of the data for training purposes, reserving the remaining 20\% for model evaluation and prediction testing. This partitioning is preceded by a random shuffling of the data to eliminate any biases that could arise from the order in which the data is presented. Categorical features are subjected to label encoding to transform these categorical values into a numerical representation. Our methodology employs a variety of advanced machine learning algorithms like Gradient Boosted Decision Trees, Random Forest, LightGBM, XGBoost, and Decision Tree models, which show superior performance in traffic incident duration prediction in recent studies \cite{grigorev2022incident,shafiei2020short}. we employ sophisticated feature importance techniques like the SHAP method, allowing us to gain insights into how individual features influence the outcome of our predictions. Feature importance estimates poses relevance to the stakeholders, highlighting which information is relevant for accurate incident duration prediction models.

\section*{Incident Duration Prediction Task}

The goal of the incident duration prediction task is to forecast the amount of time (duration) an incident will last, given a set of features related to the incident. This task can be formulated as a regression problem where the target variable is the incident duration, measured in minutes.

\begin{itemize}
    \item Let $X = \{x_1, x_2, \dots, x_n\}$ be the feature set for an incident, where each $x_i$ represents a specific attribute such as time of day, location details, road type, etc.
    \item Let $y$ represent the actual duration of an incident.
    \item The goal is to learn a function $f: X \to Y$ that maps the input features $X$ to a predicted incident duration $\hat{y}$, where $Y$ is the set of possible durations.
\end{itemize}

Another objective is to classify the incidents into two categories based on their duration:

\begin{itemize}
    \item Short-term: Incidents lasting less than or equal to the threshold.
    \item Long-term: Incidents lasting more than the threshold.
\end{itemize}

We aim to learn a function $f: \mathbb{X} \rightarrow \mathbb{Y}$, where $\mathbb{X}$ is the input space of features related to an incident, and $\mathbb{Y} = \{0, 1\}$ is the binary output space. Here, $0$ represents short-term incidents, and $1$ represents long-term incidents. For a given threshold $\tau$, the duration $d$ of an incident is classified as follows:

\begin{equation}
f(d) = 
\begin{cases} 
0 & \text{if } d \leq \tau \\
1 & \text{if } d > \tau
\end{cases}
\end{equation}

To achieve this classification, we can employ various machine learning algorithms. Each incident is represented by a set of features $X = \{x_1, x_2, \dots, x_n\}$, and the algorithm is trained to differentiate between the two classes based on these features. Performance is evaluated using metrics suitable for binary classification, such as accuracy and F1-score.

\subsection*{Evaluation Metrics}

\begin{enumerate}
    \item \textbf{Root Mean Square Error (RMSE)} measures the differences between the values predicted by a model and the values observed. It is defined as:
    \[RMSE = \sqrt{\frac{1}{n}\sum_{i=1}^{n}(\hat{y}_i - y_i)^2}\]
    where $\hat{y}_i$ is the predicted value, $y_i$ is the actual value, and $n$ is the number of samples.
    
    \item \textbf{F1 Score} is used to measure a model's accuracy on a dataset for classification tasks. It is the harmonic mean of precision and recall:
    \[F1 = 2 \times \frac{precision \times recall}{precision + recall}\]
    where $precision = \frac{TP}{TP + FP}$ and $recall = \frac{TP}{TP + FN}$.
\end{enumerate}

\subsection*{Machine Learning Methods}
\begin{enumerate}
    \item \textbf{Gradient Boosted Decision Trees (GBDT)} \cite{friedman2001greedy}: This is an ensemble technique that builds multiple decision trees in a sequential manner, where each tree corrects errors made by the previous ones. The final prediction is a weighted sum of the predictions from all trees.
    
    \item \textbf{Random Forest} \cite{breiman2001random}: An ensemble learning method that constructs a multitude of decision trees during training time and outputs the mean/median prediction of the individual trees for regression tasks, or the mode of the classes for classification tasks.
    
    \item \textbf{LightGBM} \cite{ke2017lightgbm}: Light Gradient Boosting Machine, a highly efficient gradient boosting framework that uses tree-based learning algorithms. It is designed to be distributed and efficient with faster training speed and higher efficiency, lower memory usage, and better accuracy.
    
    \item \textbf{XGBoost} \cite{chen2016xgboost}: Extreme Gradient Boosting, this is an optimized distributed gradient boosting library designed to be highly efficient, flexible, and portable. It implements machine learning algorithms under the Gradient Boosting framework, with a focus on computational speed and model performance.
    
    \item \textbf{Decision Tree} \cite{quinlan2014c4}: A decision support tool that uses a tree-like model of decisions and their possible consequences. It's a simple prediction model that maps observations about an item to conclusions about the item's target value.
\end{enumerate}

\subsection*{Feature Importance Estimation}

\begin{enumerate}
    \item \textbf{Feature Importance based on Decision Tree splits}: when using Decision Trees we can calculate the importance of features by measuring how much each feature decreases the impurity of a split (for classification) or the variation (for regression) \cite{verikas2011mining, zhou2021unbiased}. Features that are selected more often for splitting at the top of the trees are considered more important.
    
    \item \textbf{SHAP (SHapley Additive exPlanations)} \cite{lundberg2017unified}: SHAP values provide a way to explain the output of any machine learning model. It's based on the game theory concept of Shapley values. SHAP values indicate how much each feature contributes, positively or negatively, to the target variable, compared to the prediction for a baseline. This method helps in understanding the impact of each feature on the model's output across the dataset.
\end{enumerate}

\section{Results}


For the classification and regression scenarios we report the score of 5-fold cross-validation, where 80\% of data used for training each model and the rest 20\% of the data used for prediction. The data is randomly shuffled before the modelling, allowing to focus on incident characteristics rather than a timeline of incident records, providing a robust estimate of model generalizability.

The figures provided represent accuracy and F1-scores at different time thresholds for predicting if a traffic incident will be short-term or long-term. Each prediction step categorizes incidents into four parts based on Actual and Predicted durations:

True Positive (TP): Traffic incidents correctly predicted as short-term (upper-left quadrant).
False Negative (FN): Traffic incidents wrongly predicted as long-term but were actually short-term (upper-right quadrant).
False Positive (FP): Traffic incidents wrongly predicted as short-term but were actually long-term (lower-left quadrant).
True Negative (TN): Traffic incidents correctly predicted as long-term (lower-right quadrant).
Precision is the ratio of correctly predicted positive observations to the total predicted positive observations. For example, precision would be the ratio of correctly identified short-term incidents to the total number of incidents classified as short-term by the model.
Recall is the ratio of correctly predicted positive observations to all the observations in the actual positive class. For example, recall would be the ratio of correctly identified short-term incidents to the total number of actual short-term incidents, high recall would also mean that the model correctly identifies most of the actual short-term incidents.

Accuracy is the ratio of correctly predicted instances to the total instances. It is a common metric for evaluating classification models:

\[
\text{Accuracy} = \frac{TP + TN}{TP + TN + FP + FN}
\]

The F1-score is the harmonic mean of precision and recall and is often preferred over accuracy, especially in situations where the class distribution is imbalanced. It is used to balance the trade-off between precision and recall, providing a single metric to evaluate the performance of a classification model:

\[
F_1 = 2 \cdot \frac{\text{Precision} \cdot \text{Recall}}{\text{Precision} + \text{Recall}}
\]

Or in terms of the basic four elements of the confusion matrix:
\[
F_1 = 2 \cdot \frac{TP}{2TP + FP + FN}
\]

\begin{figure}[h]
    \centering
    \includegraphics[width=0.98\textwidth]{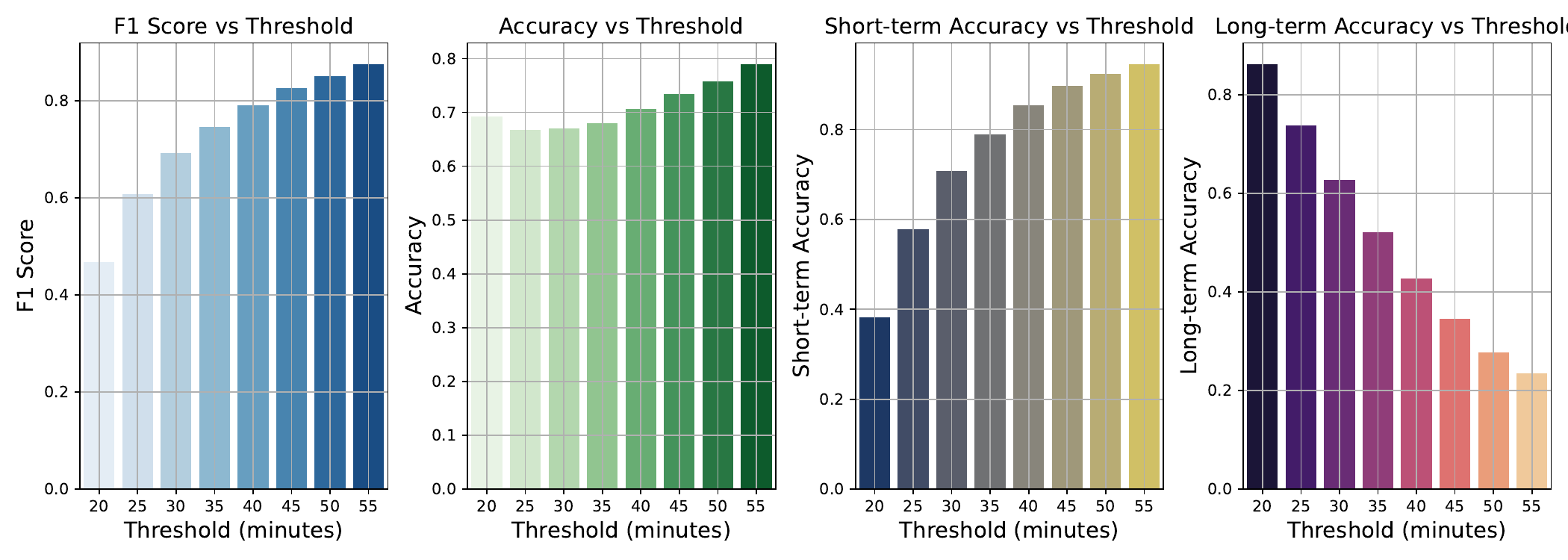}
    \caption{Barplots for F1-score, Accuracy and class balance}
    \label{fig:TIPA}
\end{figure}

As the threshold increases, model performance in classifying incidents as short-term improves significantly (as evidenced by higher F1 Scores and accuracies), but at the expense of its ability to correctly identify long-term incidents (as seen from the falling long-term accuracy) - see Figure \ref{fig:TIPA}. The model's increasing bias towards short-term predictions might lead to operational challenges, particularly in scenarios where predicting long-term incidents is crucial. Adjustments in model training or threshold settings might be required to achieve balanced prediction for both classes.

\begin{align*}
\text{Threshold 20:} &\quad \text{Accuracy = 69.29\%} \\
&\quad \text{Precision = 60.10\%} \\
&\quad \text{Recall = 38.26\%} \\
&\quad \text{F1 Score = 46.76\%} \\
&\quad \text{Short-term Accuracy = 38.26\%} \\
&\quad \text{Long-term Accuracy = 86.18\%}
\end{align*}

Here, despite a reasonable overall accuracy, the model demonstrates a low recall and short-term accuracy, suggesting it struggles to detect most of the actual short-term incidents at this threshold, while it performs better in identifying long-term incidents.

As thresholds increase progressively up to 55, observations are:

\begin{align*}
\text{Threshold 55:} &\quad \text{Accuracy = 78.96\%} \\
&\quad \text{Precision = 81.46\%} \\
&\quad \text{Recall = 94.56\%} \\
&\quad \text{F1 Score = 87.52\%} \\
&\quad \text{Short-term Accuracy = 94.56\%} \\
&\quad \text{Long-term Accuracy = 23.56\%}
\end{align*}

At this highest threshold, the metrics show high precision, recall, and F1 score for short-term incidents, indicating strong model performance in identifying and correctly labeling short-term incidents. However, there is a significant drop in long-term accuracy, highlighting a model bias towards short-term predictions and suggesting its limited capability in accurately detecting longer incident durations.

Taking a closer look at the metrics across different thresholds, the \textbf{30 minutes threshold} stands out as a balanced choice for several reasons:

\begin{enumerate}
    \item While there are higher scores in some metrics at larger thresholds, the \textbf{30 minutes threshold} offers a reasonable balance between short-term and long-term accuracy. This threshold achieves a respectable balance of \textbf{70.84\%} short-term accuracy and \textbf{62.72\%} long-term accuracy, which are superior compared to lower thresholds like 20 minutes and 25 minutes.
    \item The \textbf{Precision, Recall, and F1 Score} at this threshold are robust (\textbf{67.57\%, 70.84\%,} and \textbf{69.17\%} respectively), suggesting the model performs reliably without significant bias towards either class.
    \item Post the 30 minutes mark, although metrics like precision and recall keep increasing for short-term predictions, long-term accuracy declines sharply, indicating a strong bias towards short-term predictions in higher thresholds and potential underestimation of longer incidents.
\end{enumerate}

Based on the analysis, setting the threshold at 30 minutes provides a sustainable balance between detecting short-term and long-term traffic incidents effectively and accurately. This threshold helps in maintaining a relatively high detection rate for both short-term and long-term incidents, which is crucial for efficient traffic management.

\begin{figure}[h!]
    \centering
    \begin{subfigure}[b]{0.7\textwidth}
        \centering
        \includegraphics[width=\textwidth]{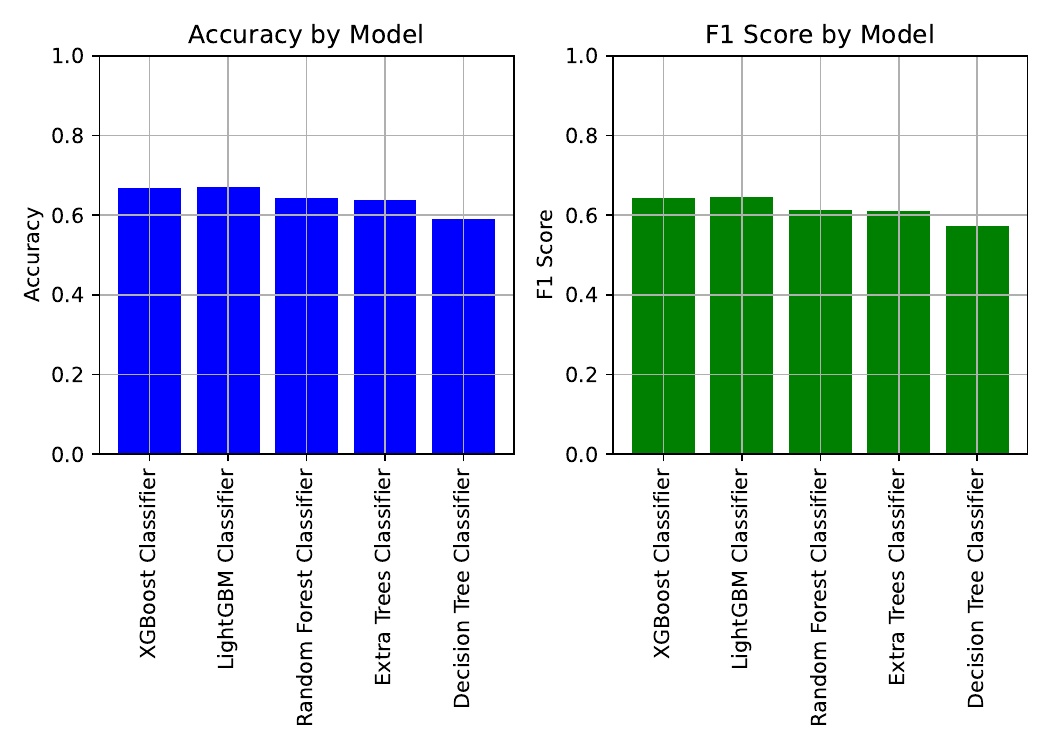}
        \caption{Comparative performance of classification models for selected duration threshold of 30 minutes.}
        \label{fig:Cmodelperformance}
    \end{subfigure}
    \vfill
    \begin{subfigure}[b]{0.7\textwidth}
        \centering
        \includegraphics[width=\textwidth]{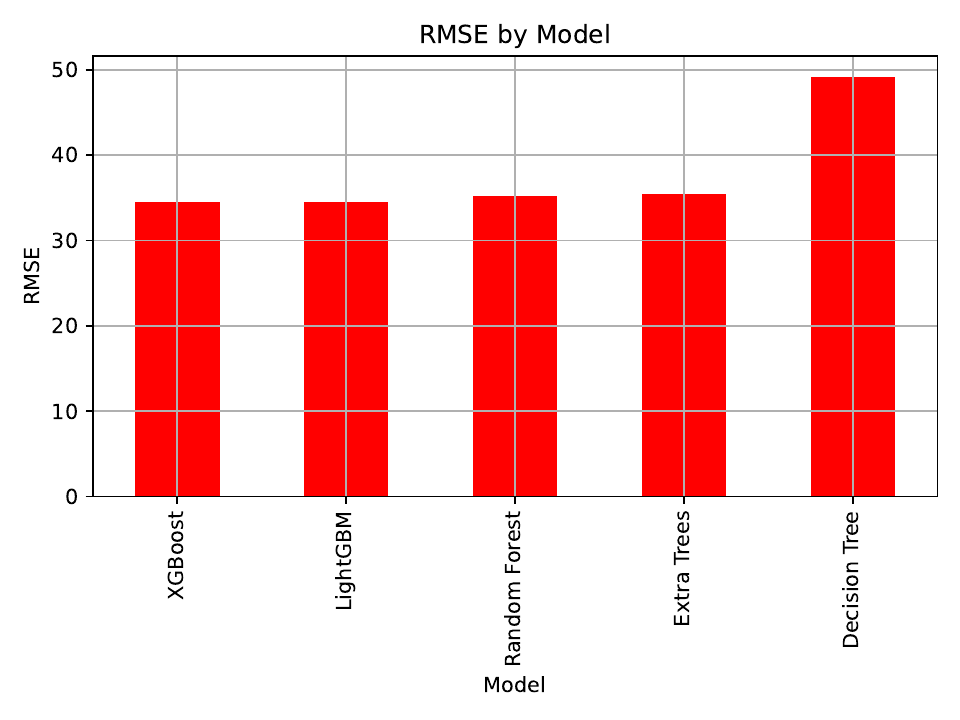}
        \caption{Comparison of model RMSE values.}
        \label{fig:Rmodelperformance}
    \end{subfigure}
    \caption{Comparative performance of models: (a) Classification metrics for a 30-minute duration threshold; (b) Model RMSE values.}
    \label{fig:combinedmodelperformance}
\end{figure}

In cases of traffic incident duration classification performance 
 (see Figure \ref{fig:Cmodelperformance}), XGBoost and LightGBM show the highest Accuracy and F1-score (0.67 and 0.62 correspondingly). The same performance is observed for regression task (see Figure \ref{fig:Rmodelperformance}, RMSE is 33.7 for XGBoost).

\clearpage

\section{Feature Importance Analysis}

For the task of feature importance analysis, we use two approaches - using feature importance obtained directly from XGboost tree splits and SHAP.
XGBoost shows number of affected lanes and 38\_PD, scale of incident, traffic volume and motorway length to be the top 5 features. Similar results were obtained when using SHAP method - types of primary and secondary vehicles were found to be the most important, while traffic volume and number of affected lanes also occupying the top 5 features.

\begin{figure}[h]
    \centering
    \begin{subfigure}[b]{0.45\textwidth}
        \includegraphics[height=7cm, width=\textwidth, keepaspectratio]{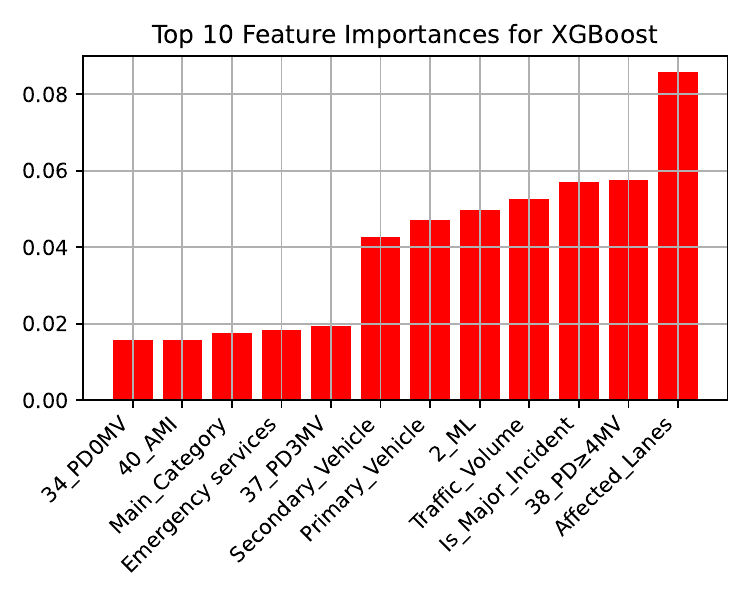}
        \caption{XGBoost model}
        \label{fig:feature_importance_xgboost}
    \end{subfigure}
    \caption{Top 10 feature importances for the XGBoost model.}
    \label{fig:feature_importances}
\end{figure}

The XGBoost model identified the top five features based on tree splits as the number of affected lanes, Feature 38\_PD4MV (Percentage of Dwellings with Four or
More Motor Vehicles), the scale of the incident, traffic volume, and motorway length (see Figure \ref{fig:feature_importances}). This indicates that the model heavily relied on these features to split the data and make predictions. The methodology involves counting the number of times a feature contributes to a decision point in the trees and evaluating the weight or gain associated with each split. This straightforward method leverages the internal structure of XGBoost trees, reflecting how often a feature is used in decision splits. It highlights features that frequently contribute to decision boundaries, focusing on the feature's role in reducing impurity or improving the criterion, such as gain or information gain, at each split.

\begin{figure}[h]
    \centering
    \includegraphics[width=0.6\textwidth]{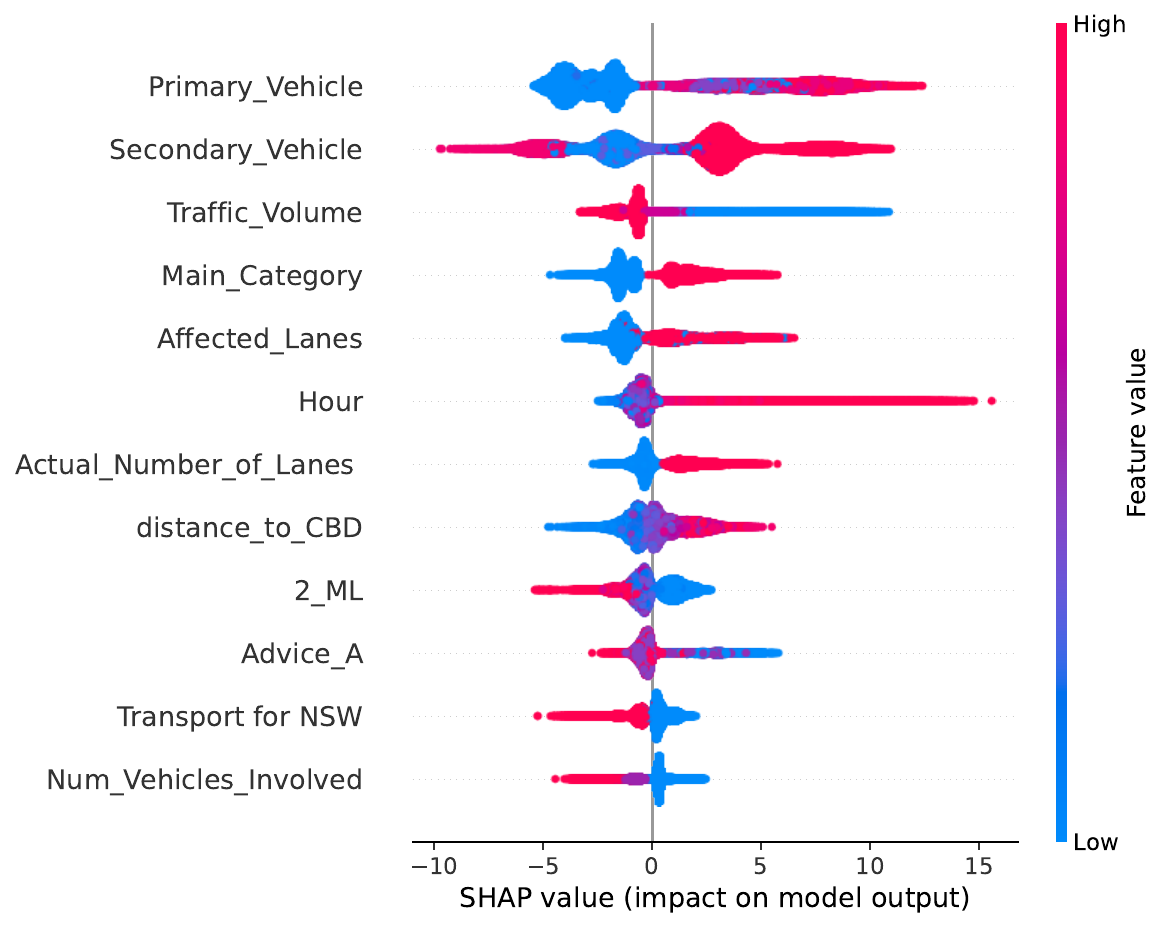}
    \caption{Top 12 feature importances as determined by SHAP values from an XGBoost model.}
    \label{fig:shap_importances}
\end{figure}

In contrast, the SHAP analysis identified the top features as the types of primary and secondary vehicles, traffic volume, and the number of affected lanes (see Figure \ref{fig:shap_importances}).

While the analysis from SHAP identified types of primary and secondary vehicles as the most crucial, features like traffic volume and number of affected lanes remained significant, thus partially aligning with the XGBoost's direct feature importance:
\begin{itemize}
    \item SHAP values provide a global explanation adjusted from local explanations. They quantify the impact of each feature on the model’s output across different predictions.
    \item SHAP values offer a sophisticated view as they consider feature interactions and the context in which each feature acts, providing comprehensive insights into how each feature contributes to predictions for individual instances and overall.
    \item They possess a solid theoretical foundation based on cooperative game theory, ensuring that contributions are fair and additive.
\end{itemize}

The differences in outcomes between the two methods can be attributed to the following reasons:
\begin{enumerate}
    \item \textbf{Perspective}: XGBoost’s tree split importance focuses on the frequency and gain related to feature splits, often isolating feature contributions in a univariate manner. In contrast, SHAP values consider the impact jointly with other features, capturing more complex interactions.
    \item \textbf{Interpretation Scope}: XGBoost importance reveals how often features drive splits in the constructed trees, influenced by tree structure and split criteria. SHAP values interpret how much predictions change with feature alterations, offering a more direct measure of feature contribution to the specific outcomes.
    \item \textbf{Feature Interactions}: SHAP explicitly quantifies interactions between features and their combined impact on predictions, which is not directly discernible from tree split counts alone. Thus, SHAP may uncover critical features not evident in univariate split importance.
\end{enumerate}

\section{Conclusion}

This study aimed to predict the duration of traffic incidents, classifying them as short-term or long-term using machine learning models, and to analyze the feature importance contributing to these predictions. The primary models utilized were Decision Trees, XGBoost, and LightGBM.

The analysis revealed several key insights regarding model performance and feature importance. For the classification task, XGBoost and LightGBM showed superior performance, with Accuracy and F1-Score values of 0.67 and 0.62 respectively for a selected threshold of 30 minutes. This threshold was found to balance performance between detecting short-term and long-term incidents effectively, achieving 70.84\% short-term accuracy and 62.72\% long-term accuracy. As the threshold increased (e.g., to 55 minutes), the models exhibited an increasing bias towards predicting short-term incidents, reducing the accuracy of long-term predictions. This indicates that careful selection of the threshold is crucial for maintaining a balanced model performance.

In the regression task, predicting the exact duration of incidents, XGBoost demonstrated the lowest RMSE value of 33.7, indicating the highest predictive accuracy among the models evaluated.

The feature importance analysis relied on two distinct approaches: feature importance based on tree splits and SHAP values. The XGBoost model identified key features influencing predictions as the number of affected lanes, the percentage of dwellings with four or more motor vehicles (Feature 38\_ PD4MV), the scale of the incident, traffic volume, and motorway length. These features were frequently used in decision-making splits within the XGBoost trees.

On the other hand, SHAP analysis provided a nuanced understanding by capturing the impact of features through their contributions to the model's output across different predictions. The SHAP values identified the types of primary and secondary vehicles, traffic volume, and the number of affected lanes as crucial features. This method emphasizes the impact of feature interactions and their contextual importance, offering a comprehensive view of how each feature contributes to the model's predictions.

The results from both the tree splits and SHAP analysis validate the significance of traffic volume, the number of affected lanes, and related features in predicting traffic incident durations. This study underscores the importance of feature selection and threshold setting in improving model accuracy and achieving balanced prediction performance for traffic incident duration classification.

Future research can utilize the same data but focus on spatial-temporal analysis \cite{articleobaid} (including hotspot detection and cluster analysis) or survival models \cite{articlesouza}.

In conclusion, the findings demonstrate the effectiveness of machine learning models in predicting traffic incident durations, with XGBoost and LightGBM offering robust performance for classification. The feature importance analysis provides valuable insights into the feature importance, highlighting the critical local area factors influencing incident duration predictions. These insights can guide future efforts in developing more accurate and balanced traffic incident management systems.

\subsection*{Acknowledgements}

This research is supported by the Australia Research Council (ARC), under scheme of the Linkage Projects (Grant ID: LP180100114).

\subsection*{CRediT author statement}
Detailed author contribution in preparation of the manuscript is described as follows.
Artur Grigorev: Conceptualization, Writing - Original Draft, Methodology, Software, Formal analysis, Investigation, Data Curation. Sajjad Shafiei: Writing - Original Draft, Writing - Review \& Editing, Conceptualization, Investigation. Adriana-Simona Mihăiță: Supervision, Methodology, Project administration. Hanna Grzybowska: Supervision, Project administration. 
\bibliographystyle{cas-model2-names}
\bibliography{bib}
\end{document}